\documentclass[11pt]{article}

\usepackage[final]{acl}

\usepackage{times}
\usepackage{latexsym}

\usepackage[T1]{fontenc}

\usepackage[utf8]{inputenc}

\usepackage{microtype}

\usepackage{inconsolata}

\usepackage{graphicx}
\usepackage{natbib}  
\usepackage{amsmath}
\usepackage{amsthm}

\usepackage{pifont}
\usepackage{afterpage}
\usepackage{placeins}

\usepackage{colortbl}
\usepackage{tabularx}
\definecolor{grpOurs}{RGB}{234,240,249}
\definecolor{grpGPT}{RGB}{252,239,226}
\definecolor{grpBase}{RGB}{241,242,243}
\definecolor{grpSFT}{RGB}{239,248,233}
\definecolor{grpGRPO}{RGB}{231,247,247}
\definecolor{grpOPD}{RGB}{245,238,250}
\usepackage{booktabs} 
\usepackage{multirow}
\usepackage{tabularray}
\usepackage{caption} 
\usepackage{algorithm}
\usepackage{algorithmic}

\usepackage{newfloat}
\usepackage{listings}

\title{CataOPD: Catalytic On-Policy Distillation for LLM Reasoning}

\author{
  Wenjin Liu\textsuperscript{1,2},
  Chenxi Wang\textsuperscript{2},
  Jiapu Wang\textsuperscript{3},
  Zhe Cui\textsuperscript{2},
  Anh Tuan Luu\textsuperscript{1},
  Haoran Luo\textsuperscript{1}\\
  \textsuperscript{1}Nanyang Technological University\\
  \textsuperscript{2}Hithink Research\\
  \textsuperscript{3}Nanjing University of Science and Technology
}

\begin{document}

\maketitle

\begin{abstract}

Reinforcement learning (RL) and on-policy distillation (OPD) enhance the reasoning capabilities of large language models (LLMs). However, when no correct trajectory is sampled, RL lacks a positive correctness signal, while OPD is constrained by trajectories reachable under the student's on-policy distribution. Therefore, we propose \textbf{CataOPD}, which uses the teacher as a catalyst rather than a target, expanding reachability and internalizing verified student-produced trajectories into a catalyst-free policy. Specifically, Self-Rescue Routing first adds on-policy samples for empirically all-failed groups, followed by Catalytic-Guided Self-Resolution for unresolved problems. Barrier-Weighted Internalization uses per-token guided-to-unaided log-probability gaps to reinforce key tokens difficult to generate without guidance. Experimental results show that \textbf{CataOPD} outperforms current baselines, improving independent solving of unresolved problems and generalization.
Our project is available at \url{https://github.com/QwenQKing/CataOPD}.
\end{abstract}

\section{Introduction}

Reasoning enables LLMs to move beyond memorizing patterns in training data and generalize to previously unseen problems~\cite{li2026rethinking}, making it a key ingredient of artificial general intelligence~\cite{wang2026don}. Through large-scale pretraining and further post-training optimization, models such as GPT-5~\cite{singh2025openai} and DeepSeek-V4~\cite{xu2026deepseek} achieve near-expert performance on demanding reasoning tasks such as mathematics~\cite{guo2025deepseek}. However, these gains remain largely confined to tasks that models can already solve reliably under their own sampling distributions~\cite{chen2026does}. Enabling models to solve currently unresolved problems beyond the reach of their unaided sampling policies remains a significant challenge~\cite{jiang2026tapo}.

\begin{figure}[t]
\centering
\includegraphics[width=0.476\textwidth]{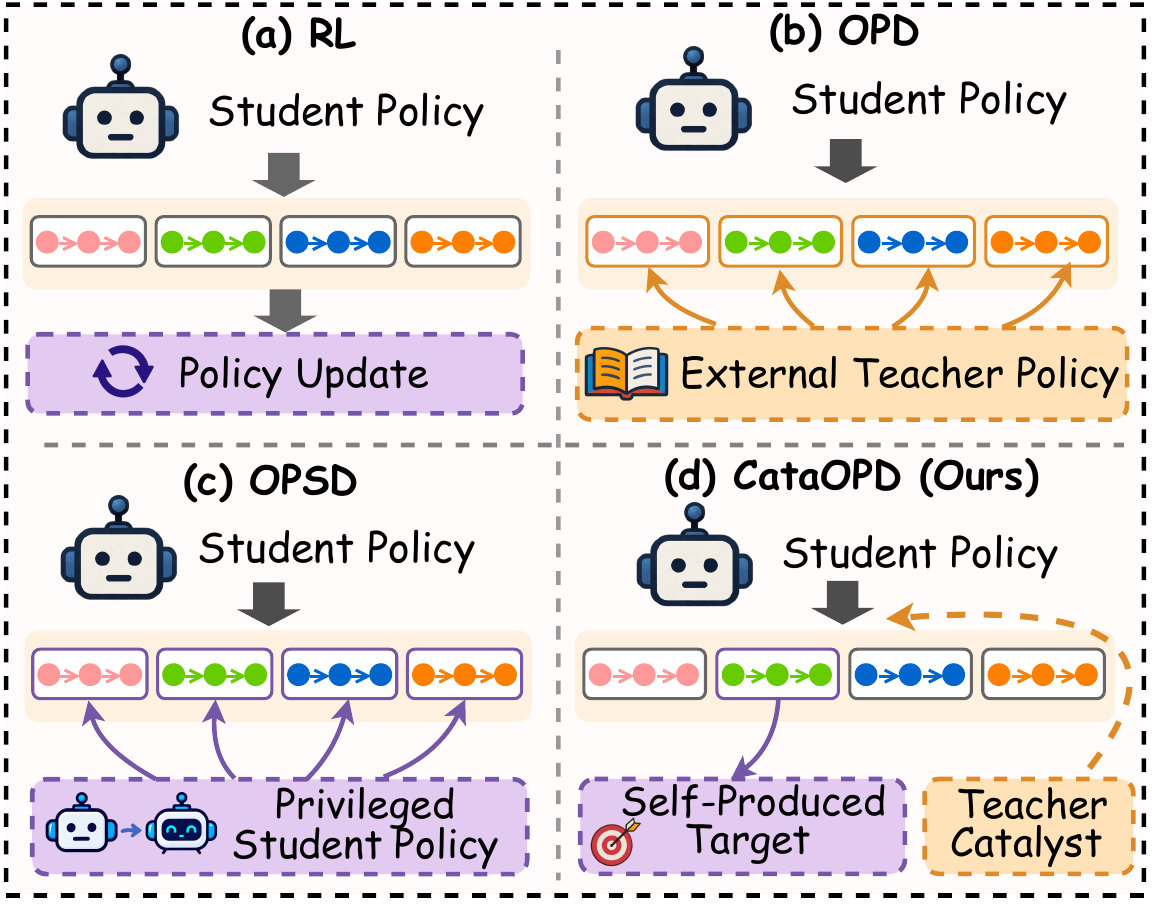} 
\caption{Comparison of CataOPD and baselines.}
\label{fig1: illustration}
\end{figure}

\begin{figure*}[t]
\centering
\includegraphics[width=1\textwidth]{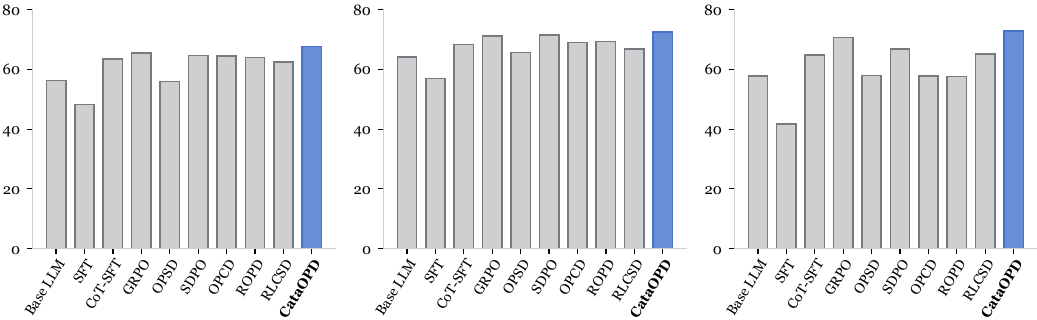}
\begin{minipage}[t]{0.32\textwidth}
\centering\small (a) Qwen2.5-3B
\end{minipage}\hfill
\begin{minipage}[t]{0.32\textwidth}
\centering\small (b) Qwen2.5-7B
\end{minipage}\hfill
\begin{minipage}[t]{0.32\textwidth}
\centering\small (c) Qwen3-1.7B
\end{minipage}
\caption{Comparison results of CataOPD and the base LLM, SFT, RL, and OPD baselines across twelve benchmarks.}
\label{fig2: related works}
\end{figure*}

RL and OPD are two methods for improving LLM reasoning~\cite{wu2026lightning}. With verifiable rewards, RL reinforces correct trajectories sampled by the model and accepted by a verifier, converting sparse outcome feedback into policy updates, such as DAPO~\cite{yu2026dapo}, GSPO~\cite{zheng2025group}, and GCPO~\cite{gu2026group}. OPD provides denser token-level teacher supervision on student-generated trajectories rather than only sparse final-outcome feedback. The teacher can be a stronger external model, including TSD-KD~\cite{kim2026explain}, HPD~\cite{zhu2026hybrid}, and Speculative KD~\cite{xu2025speculative}, or the model itself through on-policy self-distillation (OPSD), as in Self-Distilled Reasoner~\cite{zhao2026self}, Skill-SD~\cite{wang2026skill}, and OPID~\cite{yang2026opid}.

However, these methods still face three challenges: \textit{(i)} \textbf{Correctness-Signal Collapse.} When all sampled trajectories fail, group-relative RL lacks positive correctness signals and verified targets to reinforce. Their identical correctness outcomes provide no correctness-directed update for unresolved problems. \textit{(ii)} \textbf{Off-Support Distillation Target.} Correct teacher reasoning may remain unreachable under the student's unaided sampling budget. On-policy supervision can miss these trajectories, while directly distilling them introduces targets that are difficult to internalize. \textit{(iii)} \textbf{Privilege-Transfer Gap.} Privileged guidance can enable correct reasoning that remains context-dependent. Uniform token-level distillation can spread supervision over easy tokens and underweight decisive positions, weakening transfer to the student's unaided policy.

To address these challenges, we propose \textbf{CataOPD} (see Figure~\ref{fig1: illustration}), using the teacher as a catalyst rather than a target to internalize verified student-produced trajectories into a catalyst-free policy. First, \textbf{Self-Rescue Routing} addresses Correctness-Signal Collapse through additional on-policy sampling for all-failed groups, recovering verified trajectories before routing unresolved questions to catalysis. Furthermore, \textbf{Catalytic-Guided Self-Resolution} avoids Off-Support Distillation Targets by eliciting verified trajectories from the guided student distribution, providing student-produced targets rather than imitating teacher trajectories. In addition, \textbf{Barrier-Weighted Internalization} mitigates the Privilege-Transfer Gap using weights derived from clipped guided-to-unaided log-probability gaps, focusing unaided likelihood updates on key tokens difficult without guidance.

Experiments on mathematical-reasoning benchmarks show that CataOPD surpasses recent baselines (see Figure~\ref{fig2: related works}) and improves out-of-distribution generalization without a teacher at inference. Across two backbones, gains concentrate on hard problems while largely preserving base-model solvability. The trained student independently solves previously unresolved problems by internalizing its own verified trajectories, recovered through self-rescue or elicited by catalytic guidance, rather than imitating teacher trajectories.

\section{Related Work}

\textbf{Reinforcement Learning with Verifiable Rewards.}
RL with verifiable rewards~\cite{guo2025deepseek,yan2026gdepo} uses rule-based outcome checks. Group-relative methods replace value critics with normalized group rewards~\cite{song2026plan,dong2026toward}. DAPO~\cite{yu2026dapo} separates clipping bounds and adds sampling; GSPO~\cite{zheng2025group} uses sequence-level ratios; Dr. GRPO~\cite{liu2025understanding} removes length and difficulty biases. GiGPO~\cite{feng2026group} and ARPO~\cite{dong2025agentic} assign step-level credit to multi-turn agents. AAPO~\cite{xiong2026aapo} and GCPO~\cite{gu2026group} refine advantages through momentum and causal structure. ProRL~\cite{liu2026prorl} extends training with reference-policy resets, while LUFFY~\cite{yan2026learning} and ExGRPO~\cite{zhan2025exgrpo} introduce off-policy demonstrations and replay for policy optimization.

\textbf{On-Policy Distillation.}
OPD gives student rollouts dense token-level teacher signals~\cite{xie2026llm,xu2026harnessing,liu2026multi}. External-teacher OPD uses DistiLLM-2's contrastive objective~\cite{ko2025distillm}, AlignDistil's token alignment~\cite{zhang2025aligndistil}, TIP's importance-based selection~\cite{xu2026tip}, and selective or hybrid divergences in TSD-KD~\cite{kim2026explain} and HPD~\cite{zhu2026hybrid}. OPSD uses one model as privileged teacher and unprivileged student~\cite{liu2026teacher}: Self-Distilled Reasoner~\cite{zhao2026self} and $\pi$-Distill~\cite{penaloza2026privileged} use reasoning and expert traces; Skill-SD~\cite{wang2026skill} and OPID~\cite{yang2026opid} use skills; SD-Zero~\cite{he2026self} uses reward-based revision; HDPO~\cite{ding2026hdpo} supplies answers to zero-reward prompts; DemoPSD~\cite{li2026demopsddisagreementmodulatedpolicyselfdistillation} uses disagreement-based interpolation in training.

\begin{figure*}[t]
\centering
\includegraphics[width=1\textwidth]{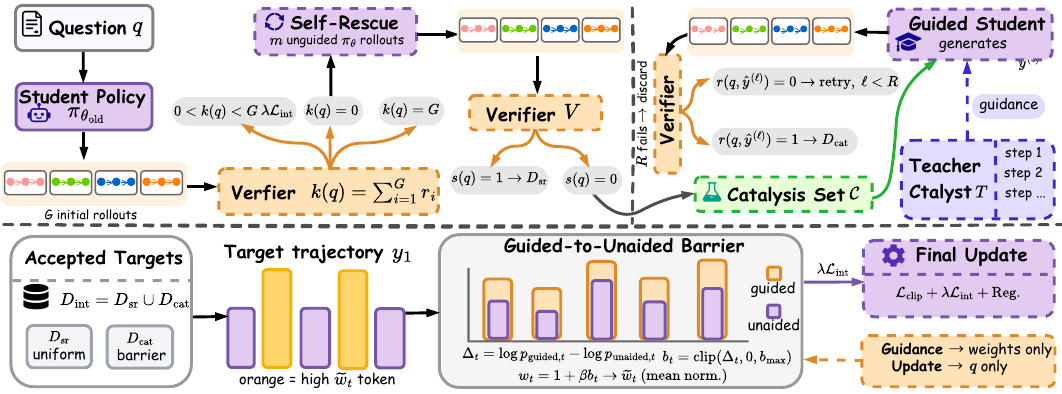} 
\caption{Overview of CataOPD, a catalytic OPD framework where the teacher acts as a catalyst rather than a target.}
\label{fig3: Bench Construction}
\end{figure*}

\section{Problem Formulation}

CataOPD uses training-time catalysis to internalize student-produced correct trajectories rather than teacher trajectories. Let $q$ be a question, $a^\star(q)$ its verification target, and $y=(y_1,\ldots,y_{|y|})$ a reasoning trajectory ending in a final answer. The student policy $\pi_\theta$ admits the autoregressive factorization
\begin{equation}
\pi_\theta(y \mid q)
=
\prod_{t=1}^{|y|}
\pi_\theta(y_t \mid q,y_{<t}),
\end{equation}
where $\theta$ denotes policy parameters, $|y|$ the trajectory length, $y_t$ the token at position $t$, and $y_{<t}=(y_1,\ldots,y_{t-1})$ the previously generated tokens.

Let $\mathrm{ans}(y)$ extract the final answer and $V$ be a rule-based verifier. We use the following binary correctness signal for both verification and routing:
\begin{equation}
r(q,y)
=
V(\mathrm{ans}(y),a^\star(q))
\in\{0,1\},
\end{equation}
where $r(q,y)$ indicates whether $V$ accepts $\mathrm{ans}(y)$. The reward $R(q,y)\in[-1,1]$ combines correctness and format rewards to compute GRPO advantages. In all-failed groups, $r(q,y)=0$, yet format rewards may differ, yielding nonzero advantages.

Define the unaided pass rate of question $q$ as the expected binary correctness of trajectories sampled from the current policy without teacher guidance:
\begin{equation}
p_0(q)
=
\mathrm{E}_{y\sim\pi_\theta(\cdot\mid q)}[r(q,y)].
\end{equation}
CataOPD uses self-rescue for all-failed groups and external catalysis for the questions still unresolved.

\section{Method}

In this section, we introduce the proposed CataOPD in detail (as shown in Figure~\ref{fig3: Bench Construction}), which includes Self-Rescue Routing, Catalytic-Guided Self-Resolution, and Barrier-Weighted Internalization.

\subsection{Self-Rescue Routing}

Self-Rescue Routing treats empirically all-failed groups as routing signals, expands self-sampling to recover a verified trajectory, and reserves external catalysis for failures unresolved by self-rescue.

For a question $q$, draw a group of $G$ trajectories from $\pi_{\theta_{\mathrm{old}}}$, a frozen snapshot of $\pi_\theta$, and compute their group-relative advantages for policy updating:
\begin{equation}
\vcenter{\hbox{$
\begin{array}{rcl@{\qquad}rcl}
\hat{A}_i
&=&
\displaystyle
\frac{R_i-\mu_q^R}{\sigma_q^R+\delta},
&
\mu_q^R
&=&
\displaystyle
\frac{1}{G}\sum_{j=1}^{G}R_j,
\\[1.2ex]
\multicolumn{6}{c}{
\displaystyle
\sigma_q^R
=
\sqrt{\frac{1}{G-1}\sum_{j=1}^{G}(R_j-\mu_q^R)^2}
}
\\[-0.2ex]
\end{array}
$}}
\end{equation}
where $i,j$ index trajectories, $R_i=R(q,y_i)$, and $\mu_q^R,\sigma_q^R$ are the group reward mean and standard deviation; $\delta>0$ prevents division by zero during advantage normalization within each group.

Equal $R_i$, not uniform correctness alone, yield $\sigma_q^R=0$ and $\hat{A}_i\equiv0$; routing instead uses the correctness count among the initial student rollouts
\begin{equation}
\begin{aligned}
k(q)&=\sum_{i=1}^{G}r_i\in\{0,\ldots,G\},\\
r_i&=r(q,y_i)\in\{0,1\},
\end{aligned}
\end{equation}
where $k(q)$ counts verified correct trajectories; $k(q)=0$ is all-failed and $k(q)=G$ all-correct. CataOPD focuses on $k(q)=0$ because no correct trajectory is available, while further student self-sampling may still recover a verified one.

For a question with $k(q)=0$, Self-Rescue Routing draws $m$ additional student trajectories for self-rescue without conditioning on teacher guidance:
\begin{equation}
\begin{aligned}
\{\tilde{y}_j\}_{j=1}^{m}
&\stackrel{\mathrm{i.i.d.}}{\sim}
\pi_{\theta_{\mathrm{old}}}(\cdot\mid q),
\\
s(q)
&=
\mathbf{1}\!\left[
\begin{gathered}
\exists j\in\{1,\ldots,m\}:\\[-0.3ex]
r(q,\tilde{y}_j)=1
\end{gathered}
\right],
\\
j^\star
&=
\min\{j:r(q,\tilde{y}_j)=1\},
\\
y^{\mathrm{sr}}
&=
\tilde{y}_{j^\star},
\qquad \mbox{if }s(q)=1,
\end{aligned}
\end{equation}
where $m$ is the self-rescue sampling budget, $j$ indexes $\tilde{y}_j$, $s(q)$ flags success, $j^\star$ the first correct index, and $y^{\mathrm{sr}}$ the selected trajectory. This unguided on-policy trajectory is the internalization target for the unaided policy used at inference on $q$ alone.

If self-rescue still fails, $q$ is routed to the catalysis set for subsequent external teacher intervention:
\begin{equation}
\mathcal{C}
=
\{q:k(q)=0\wedge s(q)=0\},
\end{equation}
where $\mathcal{C}$ contains all-failed questions unresolved by self-rescue. Catalytic-Guided Self-Resolution operates only on $\mathcal{C}$, reserving external catalysis for these still-unrecovered cases under this policy.

\noindent\textbf{Proposition 1.} \textit{For an empirically all-failed question, additional self-sampling under a fixed policy strictly increases the probability of recovering a correct trajectory whenever that policy's success probability for $q$ is strictly between zero and one.}
\vspace{-1.6mm}
\begin{proof}
We provide experimental results in Section~\ref{sec:exp_main} and theoretical proofs in Appendix~\ref{proof1}.
\end{proof}

\subsection{Catalytic-Guided Self-Resolution}

For questions unresolved by self-rescue, teacher guidance elicits verified student trajectories as training targets instead of teacher-generated solutions.

For $q\in\mathcal{C}$, select a failed initial-group trajectory $\hat{y}^{(0)}$. Over at most $H$ catalytic rounds, the external teacher $T$ produces guidance conditioned on the question, reference answer, and interaction history
\begin{equation}
\begin{aligned}
c_\ell
&=
T(q,a^\star(q),h_{\ell-1}^{T}),
\qquad 1\le \ell\le H,
\end{aligned}
\end{equation}
where $H$ is the catalytic-round limit and $h_{\ell-1}^{T}$ is the complete teacher-side history of prior guidance and student attempts. The teacher is instructed to withhold the reference answer and provide method-level guidance $c_\ell$ for the student's next attempt.

Using only the question, latest failed attempt, and current guidance, the fixed sampling policy samples a new student trajectory for verification
\begin{equation}
\begin{aligned}
\hat{y}^{(\ell)}
&\sim
\pi_{\theta_{\mathrm{old}}}(\cdot\mid q,\hat{y}^{(\ell-1)},c_\ell),
\\[-0.3ex]
&\qquad\ell=1,\ldots,H,
\end{aligned}
\end{equation}
where $\hat{y}^{(\ell)}$ is the student sample at round $\ell$. The same verifier checks each candidate; $\ell^\star$ denotes the first verified round and $y^{\mathrm{cat}}$ its accepted trajectory for internalization into the unaided student policy:
\begin{equation}
\begin{aligned}
\ell^\star
&=\!
\min\{\ell:r(q,\hat{y}^{(\ell)})=1\},
\\
y^{\mathrm{cat}}
&=\!
\hat{y}^{(\ell^\star)}
\quad
\mbox{if it exists}.
\end{aligned}
\end{equation}
If no $\ell^\star$ exists within $H$ rounds, the question is excluded from internalization. Otherwise, $y^{\mathrm{cat}}$ is a verified student-produced trajectory sampled from the guided student distribution during training. The accepted catalytic samples form $\mathcal D_{\mathrm{cat}}=\{(q,y^{\mathrm{cat}},c_{\ell^\star}):q\in\mathcal C,\ \ell^\star\text{ exists}\}$ for training.

\noindent\textbf{Proposition 2.} \textit{If any catalytic round has positive success probability, retry reaches a verified student-produced target with positive probability.}
\vspace{-3.6mm}
\begin{proof}
We provide experimental results in Section~\ref{sec:exp_main} and theoretical proofs in Appendix~\ref{proof2}.
\end{proof}

\subsection{Barrier-Weighted Internalization}

Guided-to-unaided token log-probability gaps weight correct student trajectories, focusing unaided updates on tokens dependent on guidance.

Let $y^\star\in\{y^{\mathrm{sr}},y^{\mathrm{cat}}\}$ denote an accepted student-produced correct trajectory. Set $c(q)=c_{\ell^\star}$ for catalytic samples and $c(q)=\emptyset$ for self-rescue. The guided context below omits prior failures to isolate the contribution of catalytic guidance per token:
\begin{equation}
\tilde{x}(q)=[q;c(q)].
\end{equation}
When $y^\star=y^{\mathrm{sr}}$, $\tilde{x}(q)$ reduces to $q$, so guided and unaided token probabilities coincide and barrier weighting reduces to uniform internalization for that trajectory across its valid token positions.

For token $t$ in $y^\star$, CataOPD defines the guided-to-unaided barrier that weights this token during internalization training for each accepted sample:
\begin{equation}
\vcenter{\hbox{$
\begin{array}{rcl}
b_t
&=&
\mathrm{clip}\!\left(\Delta_t,0,b_{\max}\right),
\\
\Delta_t
&=&
\mathrm{sg}\!\left[
\begin{aligned}
&
\log\pi_\theta(y_t^\star\mid \tilde{x}(q),y_{<t}^\star)
\\
&{}
-
\log\pi_\theta(y_t^\star\mid q,y_{<t}^\star)
\end{aligned}
\right],
\end{array}
$}}
\end{equation}
where $\mathrm{sg}[\cdot]$ denotes stop-gradient, $b_{\max}>0$ is the clipping threshold, and $b_t$ is the clipped gap. Large $b_t$ marks a token substantially more probable with guidance given the same prefix, while $\mathrm{sg}[\cdot]$ fixes barrier weights during differentiation of loss $\mathcal{L}_{\mathrm{int}}$.

The barrier induces token weight $w_t=1+\beta b_t\ge1$, where $\beta\ge0$ controls barrier amplification. The zero setting recovers uniform internalization, while positive settings give higher-barrier tokens larger unnormalized weights before mean normalization.

Let $m_t\in\{0,1\}$ mask valid tokens of $y^\star$, and let $t'$ index token positions. To avoid loss-scale drift, mean-normalize valid-token weights within each trajectory as follows, using the binary token mask:
\begin{equation}
\tilde{w}_t
=
w_t\cdot
\frac{\sum_{t'}m_{t'}}
{\sum_{t'}w_{t'}m_{t'}}.
\end{equation}
Accordingly, the average valid-token weight remains one, preserving each trajectory's loss scale:
\begin{equation}
\frac{\sum_t m_t\tilde{w}_t}{\sum_t m_t}=1.
\end{equation}

Finally, the unaided policy internalizes the weighted student-produced correct trajectory without guidance in its conditioning context per token:
\begin{equation}
\mathcal{L}_{\mathrm{int}}(\theta)
=
-
\frac{
\sum_t m_t\tilde{w}_t
\log\pi_\theta(y_t^\star\mid q,y_{<t}^\star)
}{
\sum_t m_t
}.
\end{equation}
Within this loss, guidance enters only through detached barrier weights, while likelihood updates condition only on $q$ and act on the unaided policy used for catalyst-free inference from question $q$.

\noindent\textbf{Proposition 3.} \textit{Enabling barrier weighting gives tokens above the trajectory-mean barrier larger direct first-order unaided update coefficients than uniform internalization on the same trajectory.} \vspace{-1.6mm}
\begin{proof}
We provide experimental results in Section~\ref{sec:exp_main} and theoretical proofs in Appendix~\ref{proof3}.
\end{proof}

\renewcommand{\dbltopfraction}{0.98}
\renewcommand{\textfraction}{0.02}
\renewcommand{\dblfloatpagefraction}{0.9}
\setcounter{dbltopnumber}{3}
\begin{table*}[t]
\centering
\begingroup
\fontsize{8.5pt}{8.7pt}\selectfont
\setlength{\tabcolsep}{0pt}
\renewcommand{\arraystretch}{1.01}
\begin{tabularx}{\textwidth}{@{}>{\raggedright\arraybackslash}p{0.14\textwidth}*{9}{>{\centering\arraybackslash}X}@{}}
\toprule
\textbf{Method} & \textbf{MetaMath} & \textbf{GSM+} & \textbf{BigMath} & \textbf{OmniMath} & \textbf{Numina} & \textbf{MATH} & \textbf{DeepMath} & \textbf{OpenMath} & \textbf{Average} \\
\midrule
\mbox{GPT-5.4-mini} & 89.84 & 88.28 & 78.12 & 42.19 & 69.53 & 85.16 & 43.75 & 41.41 & 67.29 \\
\midrule
\multicolumn{10}{c}{\textit{Qwen2.5-3B}} \\
\midrule
\mbox{Base LLM} & 70.31 & 75.00 & 59.38 & 34.38 & 59.38 & 64.06 & 52.34 & 39.06 & 56.74 \\
\mbox{SFT} & 64.84 & 32.81 & 55.47 & 24.22 & 50.00 & 53.91 & 57.03 & 35.94 & 46.78 \\
\mbox{CoT-SFT} & 84.38 & 86.72 & 71.09 & 29.69 & 71.88 & 74.22 & 58.59 & 29.69 & 63.28 \\
\mbox{GRPO} & 87.50 & 86.72 & 73.44 & 31.25 & 72.66 & 75.00 & 61.72 & 39.06 & 65.92 \\
\mbox{OPSD} & 80.47 & 82.03 & 67.19 & 26.56 & 58.59 & 63.28 & 46.88 & 28.91 & 56.74 \\
\mbox{SDPO} & 87.50 & 85.94 & 71.09 & 33.59 & 70.31 & 75.00 & 62.50 & 38.28 & 65.53 \\
\mbox{OPCD} & 85.16 & 82.03 & 71.88 & 35.16 & 71.88 & 75.78 & 62.50 & 40.62 & 65.62 \\
\mbox{ROPD} & 86.72 & 86.72 & 73.44 & 35.16 & 70.31 & 75.78 & 57.81 & 38.28 & 65.53 \\
\mbox{RLCSD} & 83.59 & 84.38 & 71.88 & 28.91 & 67.19 & 74.22 & 60.94 & 39.84 & 63.87 \\
\rowcolor{grpOurs}[0pt][0pt]\mbox{\textbf{CataOPD}} & \textbf{88.28} & \textbf{87.50} & \textbf{74.22} & \textbf{35.94} & \textbf{73.44} & \textbf{76.56} & \textbf{65.62} & \textbf{41.41} & \textbf{67.87} \\
\midrule
\multicolumn{10}{c}{\textit{Qwen2.5-7B}} \\
\midrule
\mbox{Base LLM} & 87.50 & 82.03 & 74.22 & 32.03 & 71.09 & 72.66 & 57.03 & 40.62 & 64.65 \\
\mbox{SFT} & 74.22 & 56.25 & 64.84 & 24.22 & 65.62 & 65.62 & 63.28 & 44.53 & 57.32 \\
\mbox{CoT-SFT} & 85.94 & 88.28 & 78.12 & 35.94 & 75.00 & 77.34 & 64.84 & 42.97 & 68.55 \\
\mbox{GRPO} & 89.06 & 92.19 & 78.12 & 35.94 & 76.56 & 82.81 & 69.53 & 50.78 & 71.88 \\
\mbox{OPSD} & 85.16 & 89.06 & 74.22 & 34.38 & 70.31 & 75.78 & 62.50 & 39.84 & 66.41 \\
\mbox{SDPO} & 88.28 & 92.19 & 78.12 & 35.16 & 76.56 & 82.81 & 69.53 & 50.78 & 71.68 \\
\mbox{OPCD} & 85.94 & 90.62 & 74.22 & 35.94 & 72.66 & 81.25 & 68.75 & 46.88 & 69.53 \\
\mbox{ROPD} & 89.06 & 90.62 & 78.12 & 35.94 & 75.78 & 82.81 & 57.81 & 43.75 & 69.24 \\
\mbox{RLCSD} & 84.38 & 89.84 & 74.22 & 35.94 & 70.31 & 78.12 & 66.41 & 43.75 & 67.87 \\
\rowcolor{grpOurs}[0pt][0pt]\mbox{\textbf{CataOPD}} & \textbf{89.84} & \textbf{92.97} & \textbf{78.91} & \textbf{36.72} & \textbf{77.34} & \textbf{83.59} & \textbf{70.31} & \textbf{51.56} & \textbf{72.66} \\
\midrule
\multicolumn{10}{c}{\textit{Qwen3-1.7B}} \\
\midrule
\mbox{Base LLM} & 78.12 & 87.50 & 70.31 & 27.34 & 67.97 & 61.72 & 33.59 & 17.97 & 55.57 \\
\mbox{SFT} & 54.69 & 23.44 & 51.56 & 19.53 & 39.84 & 46.88 & 56.25 & 32.81 & 40.62 \\
\mbox{CoT-SFT} & 82.81 & 83.59 & 70.31 & 38.28 & 69.53 & 73.44 & 65.62 & 43.75 & 65.92 \\
\mbox{GRPO} & 85.16 & 88.28 & 74.22 & 40.62 & 75.78 & 78.91 & 67.19 & 44.53 & 69.34 \\
\mbox{OPSD} & 82.81 & 89.06 & 68.75 & 30.47 & 66.41 & 67.97 & 25.78 & 17.97 & 56.15 \\
\mbox{SDPO} & 84.38 & 86.72 & 68.75 & 37.50 & 76.56 & 73.44 & 62.50 & 43.75 & 66.70 \\
\mbox{OPCD} & 82.81 & 85.16 & 67.97 & 29.69 & 62.50 & 68.75 & 38.28 & 19.53 & 56.84 \\
\mbox{ROPD} & 82.03 & 88.28 & 69.53 & 30.47 & 64.84 & 66.41 & 26.56 & 19.53 & 55.96 \\
\mbox{RLCSD} & 84.38 & 88.28 & 73.44 & 34.38 & 71.09 & 77.34 & 52.34 & 28.91 & 63.77 \\
\rowcolor{grpOurs}[0pt][0pt]\mbox{\textbf{CataOPD}} & \textbf{85.94} & \textbf{89.84} & \textbf{75.00} & \textbf{41.41} & \textbf{79.69} & \textbf{80.47} & \textbf{67.97} & \textbf{51.56} & \textbf{71.48} \\
\bottomrule
\end{tabularx}
\endgroup
\captionsetup{width=\textwidth,justification=justified,singlelinecheck=false}
\caption{Main results of CataOPD and baselines on eight benchmarks. Accuracy are in \%. Best values are bold.
}
\label{tab:main-id-results}
\end{table*}

\subsection{Catalytic-Guided Training}

Rollout policy optimization, weighted internalization, and reference-policy regularization jointly train the student policy for catalyst-free inference.

Barrier-Weighted Internalization uses the accepted self-rescue set $\mathcal D_{\mathrm{sr}}=\{(q,y^{\mathrm{sr}},\emptyset):k(q)=0,\ s(q)=1\}$, where $\emptyset$ denotes no guidance. The internalization training set is $\mathcal D_{\mathrm{int}}=\mathcal D_{\mathrm{sr}}\cup\mathcal D_{\mathrm{cat}}$.

For each initial group, define the clipped and dual-clipped token surrogates and current-to-old policy ratio for each sampled student token $(i,t)$
\begin{equation}
\vcenter{\hbox{$
\begin{aligned}
u_{i,t}(\theta)
&=
\min\!\left(
\rho_{i,t}\hat{A}_i,
\mathrm{clip}(\rho_{i,t},1-\epsilon,1+\epsilon)\hat{A}_i
\right),
\\
\chi_{i,t}(\theta)
&=
u_{i,t}(\theta)
+
\mathbf{1}[\hat{A}_i<0]
\left(\kappa\hat{A}_i-u_{i,t}(\theta)\right)_{+},
\\
\rho_{i,t}
&=
\displaystyle
\frac{\pi_\theta(y_{i,t}\mid q,y_{i,<t})}
{\pi_{\theta_{\mathrm{old}}}(y_{i,t}\mid q,y_{i,<t})},
\end{aligned}
$}}
\end{equation}
where $\mathbf{1}[\cdot]$ is the indicator, $(x)_{+}=\max(x,0)$, $\hat{A}_i$ comes from $R_i$, and the PPO and dual-clip thresholds satisfy \mbox{$0<\epsilon<1$} and \mbox{$\kappa>1$}. Global token-mean aggregation gives the policy loss per batch
\begin{equation}
\mathcal{L}_{\mathrm{clip}}(\theta)
=
-
\mathrm{E}_{\mathrm{tok}}
\left[
\chi_{i,t}(\theta)
\right],
\end{equation}
where $\mathrm{E}_{\mathrm{tok}}$ is the update-batch valid-token mean. Equal $R_i$ yield zero advantages, whereas $k(q)$ only determines routing to the student recovery stages.

Let $\rho^{\mathrm{ref}}_{i,t}$ be the current-to-reference ratio for frozen $\pi_{\mathrm{ref}}$. The low-variance KL estimator and final objective are defined for catalytic training as
\begin{equation*}
\begin{aligned}
\mathcal{R}_{\mathrm{ref}}(\theta)
&=
\mathrm{E}_{\mathrm{tok}}
\Bigl[
\mathrm{clip}\!\bigl(
(\rho^{\mathrm{ref}}_{i,t})^{-1}
\\
&\qquad+\log\rho^{\mathrm{ref}}_{i,t}-1,-10,10
\bigr)
\Bigr],
\end{aligned}
\end{equation*}
\begin{equation}
\begin{aligned}
\mathcal{L}_{\mathrm{CataOPD}}(\theta)
&=
\mathcal{L}_{\mathrm{clip}}(\theta)
+
c_{\mathrm{KL}}\mathcal{R}_{\mathrm{ref}}(\theta)
\\
&\quad+
\lambda
\mathrm{E}_{\mathcal{D}_{\mathrm{int}}}
\left[
\mathcal{L}_{\mathrm{int}}(\theta)
\right],
\end{aligned}
\end{equation}
where $\mathrm{E}_{\mathcal D_{\mathrm{int}}}$ averages accepted triples, $c_{\mathrm{KL}}=0.01$, and $\lambda>0$ weights internalization. Within $\mathcal{L}_{\mathrm{int}}$, guidance $c(q)$ enters only through detached barrier weights; the optimized likelihood update remains conditioned only on $q$, without teacher guidance.

\section{Experiments and Results}
\label{sec:exp_main}

In this section, we report the experimental setup, results, and analysis. We answer the following research questions (RQs):
\textbf{RQ1:} Does CataOPD outperform the baseline?
\textbf{RQ2:} How well does CataOPD generalize?
\textbf{RQ3:} Does CataOPD improve hard problems while preserving base-model solvability on previously solved problems?
\textbf{RQ4:} How does each CataOPD component contribute?
\textbf{RQ5:} How does CataOPD recover verified supervision from all-failed rollout groups during training?
\textbf{RQ6:} How does catalytic guidance affect verified-trajectory likelihood and its unaided internalization, and where do barrier weights concentrate?

\subsection{Experimental Setup}
\textbf{Datasets.} We used twelve benchmarks. Eight in-distribution datasets are MetaMath, GSM+, BigMath, OmniMath, NuminaMath, MATH, DeepMath, and OpenMath~\cite{yu2024metamath,li2024gsm,albalak2025big,gao2025omni,numina_math_datasets,hendrycks2021measuring,he2025deepmath,moshkov2025aimo}. Four held-out OOD datasets are OpenR1-Math, MAWPS, DAPO-Math, and ORCA-Math~\cite{openr1math220k,koncel2016mawps,yu2026dapo,mitra2024orca}. More details are illustrated in Appendix~\ref{app:datasets}.

\textbf{Baselines.} Our baselines include GPT-5.4-mini, SFT, GPT-5.4-mini CoT-SFT, GRPO, OPSD, SDPO, OPCD, ROPD, and RLCSD~\cite{ouyang2022training,guo2025deepseek,zhao2026self,hubotter2026reinforcement,ye2026policy,fang2026rubric,pan2026rlcsd}. More details are in Appendix~\ref{app:baselines}.

\textbf{Evaluation Metrics.} We evaluate \textbf{CataOPD} using Accuracy (Acc.), multi-sample solvability (Pass@$k$), solved-problem retention (Ret.), all-failed-group recovery (Rec.), trajectory-mean log-probability gains ($\Delta\log p$), and barrier-signal concentration. More details are shown in Appendix~\ref{app:metrics}.

\textbf{Implementation.} GPT-5.4-mini serves as the catalytic teacher in training. The students are Qwen2.5-3B, Qwen2.5-7B~\cite{hui2024qwen2}, and Qwen3-1.7B~\cite{yang2025qwen3}. All experiments run on a single server with eight 80GB NVIDIA H100 GPUs. More details are in Appendix~\ref{app:implementation}.

\renewcommand{\dbltopfraction}{0.98}
\renewcommand{\textfraction}{0.02}
\renewcommand{\dblfloatpagefraction}{0.9}

\begin{figure*}[t]
\centering
\includegraphics[pagebox=cropbox,width=\textwidth]{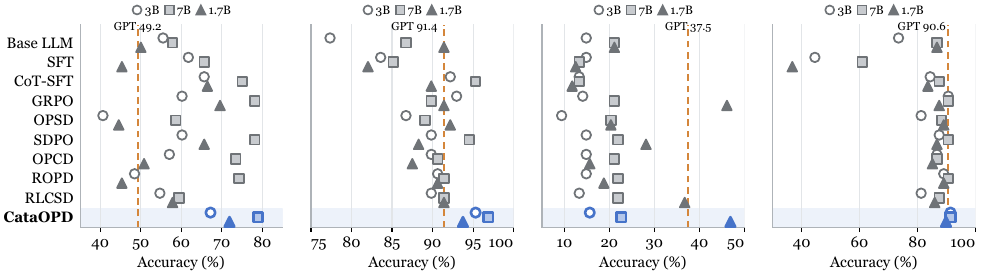}
\noindent\makebox[\textwidth][l]{
\hspace*{0.1846\textwidth}\makebox[0pt][c]{\small (a) OpenR1-Math}
\hspace*{0.2353\textwidth}\makebox[0pt][c]{\small (b) MAWPS}
\hspace*{0.2353\textwidth}\makebox[0pt][c]{\small (c) DAPO-Math}
\hspace*{0.2353\textwidth}\makebox[0pt][c]{\small (d) ORCA-Math}
}
\par\smallskip
\begin{minipage}[t]{0.345\textwidth}
\centering
\includegraphics[pagebox=cropbox,width=\linewidth]{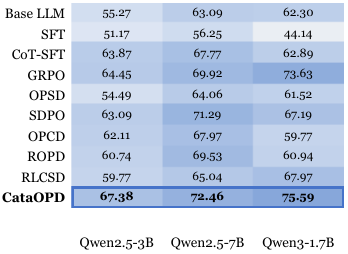}
\par\small (e) OOD average accuracy
\end{minipage}\hfill
\begin{minipage}[t]{0.278\textwidth}
\centering
\includegraphics[pagebox=cropbox,width=\linewidth]{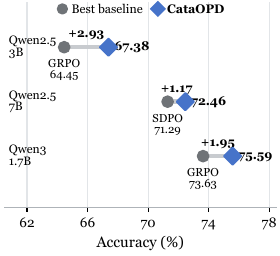}
\par\small (f) CataOPD vs. strongest baselines
\end{minipage}\hfill
\begin{minipage}[t]{0.334\textwidth}
\centering
\includegraphics[pagebox=cropbox,width=\linewidth]{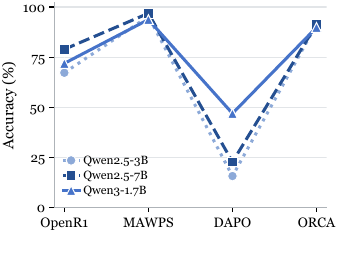}
\par\small (g) Per-benchmark OOD accuracy
\end{minipage}
\caption{Comparison of CataOPD and baselines on four OOD benchmarks. Panels (a)--(d) show per-benchmark accuracy; orange dashed lines mark GPT-5.4-mini, and blue markers/lines mark CataOPD. Panels (e)--(f) show OOD averages and margins over the strongest baseline, while (g) shows CataOPD's accuracy across benchmarks.}
\label{fig:rq2-results}
\end{figure*}

\subsection{Main Results (RQ1)}
\textbf{Overall.}
As shown in Table~\ref{tab:main-id-results}, CataOPD achieves the highest average accuracy among same-backbone methods for every student. Each variant also exceeds GPT-5.4-mini in average accuracy. These results establish an overall advantage over supervised fine-tuning, reward-based RL, and on-policy distillation baselines under catalyst-free inference on the eight evaluated benchmarks.

\textbf{Per-dataset.}
CataOPD leads all same-backbone baselines on every evaluated dataset. The largest margins over the best competing method occur on DeepMath for Qwen2.5-3B and OpenMath for Qwen3-1.7B. Positive gains elsewhere show that the advantage spans the benchmark suite.

\textbf{Per-backbone.}
Average gains over the strongest same-backbone baseline are larger for Qwen2.5-3B and Qwen3-1.7B than for Qwen2.5-7B. All three nevertheless improve, supporting effectiveness across the evaluated model families and sizes without teacher guidance at inference.

\subsection{OOD Generalization (RQ2)}

As shown in Figure~\ref{fig:rq2-results}, CataOPD's OOD results exhibit three consistent trends: \textit{(i)} \textbf{Backbone-level OOD transfer.} CataOPD achieves the highest same-backbone OOD average for all three students, supporting transfer across different model families and sizes. \textit{(ii)} \textbf{Cross-benchmark consistency.} CataOPD outperforms same-backbone baselines on all four OOD benchmarks, indicating that its advantage is not confined to one dataset. \textit{(iii)} \textbf{Comparison with the teacher.} CataOPD exceeds the GPT-5.4-mini OOD average for all three students, indicating an average OOD accuracy advantage over the teacher across all three student backbones.

\begin{figure}[t]
\centering
\includegraphics[pagebox=cropbox,width=\columnwidth]{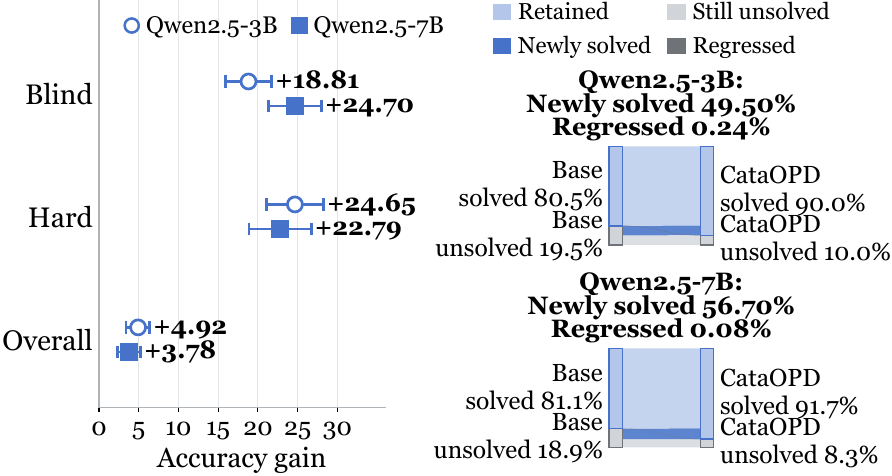}
\noindent\makebox[\columnwidth][l]{
\hspace*{0.25\columnwidth}\makebox[0pt][c]{\small (a) Gains on difficult questions}
\hspace*{0.50\columnwidth}\makebox[0pt][c]{\small (b) Solvability transitions}
}
\caption{Hard-problem analysis of CataOPD for Qwen2.5-3B and Qwen2.5-7B across all twelve evaluated benchmarks: (a) accuracy gains with 95\% confidence intervals; (b) Pass@$8$ transitions between questions solved and unsolved by the base model.}
\label{fig:rq3-hard-problem-retention}
\end{figure}

\begin{table*}[t]
\centering
\begingroup
\fontsize{8.5pt}{8.7pt}\selectfont
\setlength{\tabcolsep}{0pt}
\renewcommand{\arraystretch}{1.01}
\begin{tabularx}{\textwidth}{@{}>{\raggedright\arraybackslash}X*{8}{>{\centering\arraybackslash}X}@{}}
\toprule
\textbf{Method} & \textbf{MetaMath} & \textbf{GSM+} & \textbf{BigMath} & \textbf{OmniMath} & \textbf{Numina} & \textbf{MATH} & \textbf{DeepMath} & \textbf{OpenMath} \\
\midrule
\mbox{w/o Self-Rescue} & 62.50 & 37.50 & 53.12 & 16.41 & 44.53 & 49.22 & 45.31 & 25.00 \\
\mbox{w/o Guidance} & 87.50 & 86.72 & 71.88 & 32.81 & 72.66 & 75.78 & 64.84 & 40.62 \\
\mbox{w/o Barrier} & 81.25 & 85.94 & 72.66 & 30.47 & 67.97 & 73.44 & 63.28 & 37.50 \\
\rowcolor{grpOurs}[0pt][0pt]\textbf{CataOPD} & \textbf{88.28} & \textbf{87.50} & \textbf{74.22} & \textbf{35.94} & \textbf{73.44} & \textbf{76.56} & \textbf{65.62} & \textbf{41.41} \\
\bottomrule
\end{tabularx}
\endgroup
\captionsetup{width=\textwidth,justification=justified,singlelinecheck=false}
\caption{Component ablations for CataOPD with Qwen2.5-3B across eight benchmarks. The w/o variants remove Self-Rescue Routing, Catalytic-Guided Self-Resolution, and Barrier-Weighted Internalization, respectively.}
\label{tab:component-ablation}
\end{table*}

\begin{figure*}[!t]
\centering
\includegraphics[pagebox=cropbox,width=\textwidth]{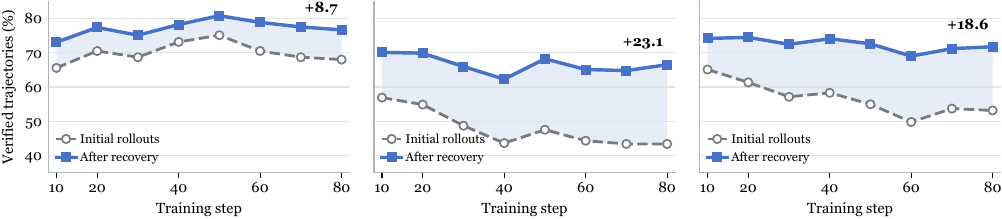}
\noindent\makebox[\textwidth][l]{
\hspace*{0.20\textwidth}\makebox[0pt][c]{\small (a) Qwen3-1.7B}
\hspace*{0.325\textwidth}\makebox[0pt][c]{\small (b) Qwen2.5-3B}
\hspace*{0.325\textwidth}\makebox[0pt][c]{\small (c) Qwen2.5-7B}
}
\caption{Training-time verified-trajectory coverage before and after recovery; annotations show final-step gaps.}
\label{fig:rq5-training-dynamics}
\vspace{3pt}
\includegraphics[pagebox=cropbox,width=\textwidth]{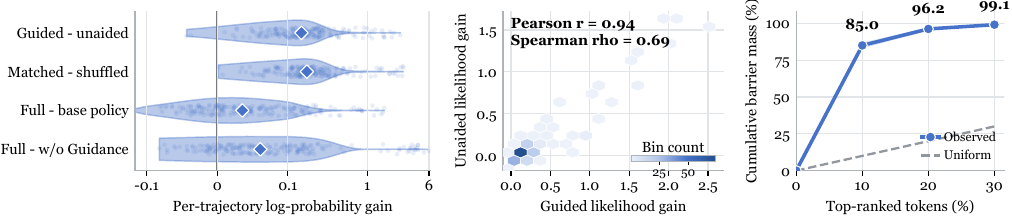}
\noindent\makebox[\textwidth][l]{
\hspace*{0.30\textwidth}\makebox[0pt][c]{\small (a) Guidance effects}
\hspace*{0.315\textwidth}\makebox[0pt][c]{\small (b) Guidance--internalization relation}
\hspace*{0.275\textwidth}\makebox[0pt][c]{\small (c) Barrier concentration}
}
\caption{Conditional mechanism diagnostics on 200 fixed trajectories recovered by catalytic guidance after initial and self-rescue failure. Full $-$ base policy and Full $-$ w/o Guidance denote trajectory-level log-probability gains over the base policy and w/o Guidance ablation under the same question-only context for each recovered trajectory.}
\label{fig:rq6-mechanisms}
\end{figure*}

\subsection{Hard-Problem Analysis (RQ3)}

Figure~\ref{fig:rq3-hard-problem-retention} analyzes difficult-question gains and retention of base-model solvability: \textit{(i)} \textbf{Hard-question accuracy gains.} For both backbones, \textit{Blind} and \textit{Hard} gains exceed \textit{Overall} gains. Larger gains on questions with no correct base-model samples or low empirical success show that improvements are not confined to questions the base model already solves reliably. \textit{(ii)} \textbf{Recovery with retained solvability.} Newly solved cases far outnumber regressions, while nearly all base-solved questions remain solvable under Pass@$8$. Since solvability requires at least one correct answer among eight samples, these transitions show an expanded solved set with limited loss of prior solvability.

\subsection{Component Ablation (RQ4)}

Table~\ref{tab:component-ablation} shows that all three CataOPD components contribute across eight benchmarks: \textit{(i)} \textbf{Self-Rescue Routing} recovers verified correct student trajectories from empirically all-failed groups, and removing it lowers every benchmark. \textit{(ii)} \textbf{Catalytic-Guided Self-Resolution} recovers additional student targets under guidance, and removing it lowers performance across the suite. \textit{(iii)} \textbf{Barrier-Weighted Internalization} prioritizes difficult tokens in recovered trajectories, and removing it also lowers performance on all eight benchmarks.

\subsection{Training Dynamics (RQ5)}

Figure~\ref{fig:rq5-training-dynamics} reports verified-trajectory coverage across all groups before and after recovery: \textit{(i)} \textbf{Recovery persists.} In every logged window, after-recovery coverage exceeds initial-rollout coverage for all three students, showing that all-failed groups continue to yield usable supervision beyond early training. \textit{(ii)} \textbf{Recovery is backbone-consistent.} The positive gap holds for Qwen3-1.7B, Qwen2.5-3B, and Qwen2.5-7B despite different initial coverage and gain magnitudes, showing that recovery remains active consistently across these student backbones throughout the full training process.

\subsection{Mechanism Diagnostics (RQ6)}

Figure~\ref{fig:rq6-mechanisms} provides conditional diagnostics on 200 Qwen2.5-3B trajectories recovered after initial and self-rescue failure: \textit{(i)} \textbf{Guidance raises verified-trajectory likelihood.} Matched guidance yields higher trajectory-level log-probability than unaided and shuffled guidance. \textit{(ii)} \textbf{Unaided likelihood improves.} Unaided internalization gain covaries positively with guided gain, consistent with selective internalization. \textit{(iii)} \textbf{Barrier mass is selective.} It concentrates on tokens with the largest guided-to-unaided log-probability gaps, focusing updates on difficult positions for targeted internalization.

\section{Conclusion}

We present \textbf{CataOPD}, a catalytic OPD framework that internalizes verified student trajectories. It combines Self-Rescue Routing, Catalytic-Guided Self-Resolution, and Barrier-Weighted Internalization. \textbf{CataOPD} achieves the best average accuracy per backbone, improves hard-problem and OOD performance, and preserves nearly all cases initially solved by the base model. Experimental results show catalytic guidance supports LLM distillation.

\newpage
\section*{Limitations}

Limitations of CataOPD mainly concern catalytic guidance and training cost. Guidance quality may vary across teachers and problem types, and the teacher may not provide equally useful signals for every unresolved question. Additional self-rescue and catalytic sampling increase training-time computation, especially for hard questions requiring repeated attempts. Efficiency and scalability under larger models and longer reasoning remain open questions for future studies.

\section*{Ethical Considerations}

We use public models and public evaluation benchmarks. The study does not involve human subjects or collect personal or sensitive information. The use of these resources follows their public licenses and stated usage terms for the experiments and released materials used in this work.

\section*{AI Assistance}

AI tools assisted code writing and language editing, while the authors designed all experiments and verified all results. Parts of Figures~1 and~3 were initially drafted with image-generation tools and then redrawn as vector graphics based on the experimental results. The experimental design, result analysis, final content, and conclusions were completed and verified by the authors for the reported experiments, figures, and conclusions reported here.

\bibliography{custom}

\appendix

\section*{Appendix}

\section{Theoretical Proof}
\label{app:theoretical-proof}

\subsection{Proof of Proposition 1}
\label{proof1}

\par\noindent\textbf{Proposition 1.} \textit{For an empirically all-failed question, additional self-sampling under a fixed policy strictly increases the probability of recovering a correct trajectory whenever that policy's success probability for $q$ is strictly between zero and one.}

\begin{proof}
Consider a question $q$ whose initial rollout group is empirically all-failed:
\begin{equation}
\vcenter{\hbox{$
\begin{array}{rcl}
y_i
&\stackrel{\mathrm{i.i.d.}}{\sim}&
\pi_{\theta_{\mathrm{old}}}(\cdot\mid q),
\qquad i=1,\ldots,G,
\\[0.8ex]
r(q,y_i)&=&0,
\qquad i=1,\ldots,G,
\end{array}
$}}
\end{equation}
where $G$ is the group size and $i$ indexes its trajectories. This event establishes that the initial group contains no verified correct trajectory. Because GRPO advantages are computed from the composite rewards $R_i$, it does not imply equal $R_i$ or zero group-relative advantages.

At the current training iteration, $\pi_{\theta_{\mathrm{old}}}$ is the fixed sampling policy. Define its unaided success probability as
\begin{equation}
p_{\mathrm{roll}}(q)
=
\Pr_{y\sim\pi_{\theta_{\mathrm{old}}}(\cdot\mid q)}
\left[r(q,y)=1\right],
\end{equation}
where the subscript $\mathrm{roll}$ identifies the policy used for rollouts. Self-Rescue Routing draws $m$ additional trajectories independently from the same policy:
\begin{equation}
\tilde y_1,\ldots,\tilde y_m
\stackrel{\mathrm{i.i.d.}}{\sim}
\pi_{\theta_{\mathrm{old}}}(\cdot\mid q),
\end{equation}
where $m$ is the self-rescue sampling budget and $j$ indexes the resulting trajectories $\tilde y_j$. At the rollout snapshot, $\pi_\theta=\pi_{\theta_{\mathrm{old}}}$, so $p_{\mathrm{roll}}(q)$ equals the unaided success probability $p_0(q)$ evaluated at that snapshot. Successful self-rescue corresponds to hitting the event $\{y:r(q,y)=1\}$. Define the first success time as
\begin{equation}
T_q
=
\min\{j\ge1:r(q,\tilde y_j)=1\}.
\end{equation}
Because the additional draws are independent of the initial group, conditioning on its observed all-failed event does not change their success probability. Their correctness indicators are therefore i.i.d. Bernoulli trials with success probability $p_{\mathrm{roll}}(q)$, and $T_q$ follows a geometric distribution:
\begin{equation}
\begin{aligned}
\Pr[T_q=j]
&=(1-p_{\mathrm{roll}}(q))^{j-1}p_{\mathrm{roll}}(q),\\
&\qquad j=1,2,\ldots.
\end{aligned}
\end{equation}
Therefore, the probability that self-rescue finds at least one correct student-produced trajectory within sampling budget $m$ is
\begin{equation}
\vcenter{\hbox{$
\begin{array}{rcl}
\Pr[T_q\le m]
&=&
\displaystyle
\sum_{j=1}^{m}(1-p_{\mathrm{roll}}(q))^{j-1}p_{\mathrm{roll}}(q)
\\[1.2ex]
&=&
\displaystyle
1-(1-p_{\mathrm{roll}}(q))^m.
\end{array}
$}}
\end{equation}
Equivalently, the self-rescue miss probability is
\begin{equation}
P_{\mathrm{miss}}^{\mathrm{sr}}(q)
=
\Pr[T_q>m]
=
(1-p_{\mathrm{roll}}(q))^m.
\end{equation}
Using $1-x\le e^{-x}$ for $x\in[0,1]$, we obtain
\begin{equation}
P_{\mathrm{miss}}^{\mathrm{sr}}(q)
\le
\exp(-mp_{\mathrm{roll}}(q)).
\end{equation}
Hence, for any question with $0<p_{\mathrm{roll}}(q)<1$, increasing the self-rescue sampling budget exponentially decreases the probability of missing every correct unaided trajectory.

When self-rescue succeeds, CataOPD obtains
\begin{equation}
y^{\mathrm{sr}}
=
\tilde y_{T_q},
\qquad
r(q,y^{\mathrm{sr}})=1.
\end{equation}
This trajectory is produced by the fixed student sampling policy without guidance. Let $m_t\in\{0,1\}$ mask its valid tokens. The self-rescue specialization of the unaided internalization loss is
\begin{equation}
\mathcal L_{\mathrm{sr}}(\theta)
=
-
\frac{
\sum_t m_t
\log\pi_\theta
(y_t^{\mathrm{sr}}\mid q,y_{<t}^{\mathrm{sr}})
}{
\sum_t m_t
},
\end{equation}
where $t$ indexes trajectory tokens and $y_{<t}^{\mathrm{sr}}$ is the prefix before position $t$. Let $z_{t,v}$ denote the softmax logit for vocabulary token $v$ at a valid position $t$. Then
\begin{equation}
\frac{\partial \mathcal L_{\mathrm{sr}}}{\partial z_{t,v}}
=
\frac{m_t}{\sum_s m_s}
\left(
\pi_\theta(v\mid q,y_{<t}^{\mathrm{sr}})
-
\mathbf 1[v=y_t^{\mathrm{sr}}]
\right).
\end{equation}
For each valid position, this logit-space gradient is nonzero unless the target token already has probability one. Thus, conditional on successful self-rescue, an empirically all-failed group obtains a correctness-directed descent direction in logit space regardless of whether its composite-reward advantages vanished.

In summary, Self-Rescue Routing performs i.i.d. Bernoulli trials whose first-success time is geometric and whose miss probability is $(1-p_{\mathrm{roll}}(q))^m\le e^{-mp_{\mathrm{roll}}(q)}$. Once reached, a correct student-produced trajectory supplies a verified target and a correctness-directed logit-space signal for unaided internalization.
\end{proof}

\subsection{Proof of Proposition 2}
\label{proof2}

\par\noindent\textbf{Proposition 2.} \textit{If any catalytic round has positive success probability, retry reaches a verified student-produced target with positive probability.}

\begin{proof}
Let
\begin{equation}
\mathcal Y^{+}(q)
=
\{y:r(q,y)=1\}
\end{equation}
be the verified correct-trajectory set for question $q$. For questions unresolved after self-rescue, CataOPD uses catalytic guidance as an input that changes the student's sampling kernel, not as a target trajectory. At round $\ell$, define
\begin{equation}
K_\ell(y\mid q,\hat y^{(\ell-1)},c_\ell)
=
\pi_{\theta_{\mathrm{old}}}(y\mid q,\hat y^{(\ell-1)},c_\ell),
\end{equation}
where $c_\ell$ is produced from the complete teacher-side history through the previous student attempt $\hat y^{(\ell-1)}$. The student then samples
\begin{equation}
\hat y^{(\ell)}
\sim
K_\ell(\cdot\mid q,\hat y^{(\ell-1)},c_\ell).
\end{equation}

This forms an $H$-round guided retry process with target set $\mathcal Y^+(q)$. Let
\begin{equation}
F_{\ell-1}
=
\bigcap_{s=1}^{\ell-1}
\{\hat y^{(s)}\notin\mathcal Y^+(q)\}
\end{equation}
denote the event that the previous $\ell-1$ attempts have not reached the target set. For reachable rounds with $\Pr[F_{\ell-1}]>0$, define the conditional success probability at round $\ell$; if $\Pr[F_{\ell-1}]=0$, set $p_\ell(q)=0$ by convention because the failure event already has zero probability:
\begin{equation}
p_\ell(q)
=
\begin{cases}
\begin{gathered}
\Pr[\hat y^{(\ell)}\in\mathcal Y^{+}(q)\\[-0.2ex]
\mid F_{\ell-1}],
\end{gathered}
& \Pr[F_{\ell-1}]>0,\\
0,
& \Pr[F_{\ell-1}]=0.
\end{cases}
\end{equation}
This probability is induced by the guided student kernel and marginalizes over the teacher-side history and catalytic guidance at round $\ell$.

For any reachable round, the chain rule gives
\(
\Pr[F_\ell]=\Pr[F_{\ell-1}](1-p_\ell(q)).
\)
If $\Pr[F_{\ell-1}]=0$, then $\Pr[F_\ell]=0$, and the same recurrence holds under the convention above. Since $\Pr[F_0]=1$, recursively applying this relation gives
\begin{equation}
\Pr[F_H]
=
\prod_{\ell=1}^{H}
(1-p_\ell(q)).
\end{equation}
Therefore, the probability of success within $H$ rounds is
\begin{equation}
\vcenter{\hbox{$
\begin{array}{rcl}
P_{\mathrm{reach}}(q)
&=&
\Pr[\exists \ell\le H:\hat y^{(\ell)}\in\mathcal Y^+(q)]
\\[0.8ex]
&=&
\displaystyle
1-
\prod_{\ell=1}^{H}(1-p_\ell(q)).
\end{array}
$}}
\end{equation}
If at least one reachable round has $p_\ell(q)>0$, then $P_{\mathrm{reach}}(q)>0$. Thus, the guided retry process has positive success probability whenever a guided student distribution assigns positive probability to the verified target set.

When a successful round exists, define
\begin{equation}
\ell^\star
=
\min\{\ell:\hat y^{(\ell)}\in\mathcal Y^+(q)\},
\qquad
y^{\mathrm{cat}}
=
\hat y^{(\ell^\star)}.
\end{equation}
Since $y^{\mathrm{cat}}$ is sampled from the guided student kernel,
\begin{equation}
K_{\ell^\star}
(y^{\mathrm{cat}}\mid q,\hat y^{(\ell^\star-1)},c_{\ell^\star})
>
0.
\end{equation}
Since it reaches the verified target set,
\begin{equation}
y^{\mathrm{cat}}\in\mathcal Y^+(q),
\qquad
r(q,y^{\mathrm{cat}})=1.
\end{equation}
Therefore, the accepted trajectory is a correct trajectory produced by the student under catalytic guidance, rather than a trajectory generated by the teacher.

CataOPD then adds $(q,y^{\mathrm{cat}},c_{\ell^\star})$ to $\mathcal D_{\mathrm{int}}$ and applies Barrier-Weighted Internalization under the unaided condition:
\begin{equation}
\begin{aligned}
\mathcal L_{\mathrm{int}}(\theta;q,y^{\mathrm{cat}},c_{\ell^\star})
&=-\frac{1}{\sum_t m_t}\sum_t m_t\tilde w_t\\[-0.2ex]
&\quad\cdot\log\pi_\theta
\!\left(y_t^{\mathrm{cat}}\mid q,y_{<t}^{\mathrm{cat}}\right).
\end{aligned}
\end{equation}
Guidance determines the weights $\tilde w_t$, but the likelihood terms condition only on $q$. Hence the learned policy remains catalyst-free at inference.

In conclusion, Catalytic-Guided Self-Resolution is an $H$-round guided retry process with target set $\mathcal Y^+(q)$. It reaches this set with probability $1-\prod_{\ell=1}^{H}(1-p_\ell(q))$, which is positive whenever some $p_\ell(q)>0$. The reached trajectory is verified and student-produced, making it a valid target for unaided internalization without teacher trajectories.
\end{proof}

\subsection{Proof of Proposition 3}
\label{proof3}

\par\noindent\textbf{Proposition 3.} \textit{Enabling barrier weighting gives tokens above the trajectory-mean barrier larger direct first-order unaided update coefficients than uniform internalization on the same trajectory.}

\begin{proof}
For an accepted trajectory $y^\star$, define the guided-to-unaided log-probability gap
\begin{equation}
\Delta_t
=
\mathrm{sg}\!\left[
\begin{aligned}
&\log\pi_\theta(y_t^\star\mid \tilde x(q),y_{<t}^\star)\\[-0.2ex]
&-\log\pi_\theta(y_t^\star\mid q,y_{<t}^\star)
\end{aligned}
\right].
\end{equation}
CataOPD converts this gap into a nonnegative barrier:
\begin{equation}
b_t
=
\mathrm{clip}(\Delta_t,0,b_{\max}).
\end{equation}
Here $\mathrm{sg}[\cdot]$ treats the gap as fixed during differentiation. The barrier measures how much guidance raises the probability of token $y_t^\star$ relative to the unaided context.

Barrier-Weighted Internalization assigns the unnormalized weight
\begin{equation}
w_t=1+\beta b_t,
\qquad
\beta\ge0.
\end{equation}
Let $B=\{t:m_t=1\}$ be the valid token set. The normalized token weight is
\begin{equation}
\tilde w_t
=
w_t
\frac{|B|}
{\sum_{s\in B}w_s}.
\end{equation}
Assuming $|B|>0$, the normalized weights satisfy
\begin{equation}
\frac{1}{|B|}
\sum_{t\in B}
\tilde w_t
=
1.
\end{equation}
Thus the weighting acts as normalized token-level reweighting while keeping the mean weight over valid tokens equal to one.

Let
\begin{equation}
\bar b
=
\frac{1}{|B|}
\sum_{t\in B}b_t
\end{equation}
be the average barrier. Since
\begin{equation}
\sum_{s\in B}w_s
=
\sum_{s\in B}(1+\beta b_s)
=
|B|(1+\beta\bar b),
\end{equation}
we obtain
\begin{equation}
\tilde w_t
=
\frac{1+\beta b_t}{1+\beta\bar b}.
\end{equation}
For $\beta>0$,
\begin{equation}
b_t>\bar b
\Rightarrow
\tilde w_t>1,
\qquad
b_t<\bar b
\Rightarrow
\tilde w_t<1.
\end{equation}
Setting $\beta=0$ makes all normalized weights equal one and recovers uniform internalization. Under barrier weighting, tokens above the trajectory-average barrier receive above-average update weight.

During internalization, the barrier weights serve as fixed token-level coefficients in the likelihood objective. Define
\begin{equation}
\ell_t(\theta)
=
\log\pi_\theta(y_t^\star\mid q,y_{<t}^\star).
\end{equation}
Let $g_t=\nabla_\theta\ell_t(\theta)$ for compactness. The internalization objective is
\begin{equation}
\mathcal L_{\mathrm{int}}(\theta)
=
-
\frac{1}{|B|}
\sum_{t\in B}
\tilde w_t
\ell_t(\theta).
\end{equation}
To isolate the contribution of $\mathcal L_{\mathrm{int}}$ within the joint training update, consider one gradient step driven by this loss:
\begin{equation}
\theta^+
=
\theta-\eta\nabla_\theta\mathcal L_{\mathrm{int}}(\theta),
\end{equation}
For locally smooth $\ell_t$ and small $\eta$, Taylor expansion gives
\begin{equation}
\begin{aligned}
\ell_t(\theta^+)-\ell_t(\theta)
&=
\displaystyle
\left\langle
g_t,
\theta^+-\theta
\right\rangle
+
O(\eta^2)
\\[1.2ex]
&=
\displaystyle
\frac{\eta}{|B|}
\tilde w_t
\|g_t\|^2
+
\frac{\eta}{|B|}
C_t
+
O(\eta^2),
\end{aligned}
\end{equation}
where the cross-token interaction is
\begin{equation}
C_t
=
\sum_{s\in B,s\ne t}
\tilde w_s
\left\langle
g_t,
g_s
\right\rangle.
\end{equation}
The first term is the direct first-order increase in the unaided log-probability of token $t$, and its coefficient scales linearly with $\tilde w_t$. Compared with uniform internalization, where $\tilde w_t=1$ for all tokens, any token with $b_t>\bar b$ receives a larger direct first-order update coefficient. The second term, $C_t$, captures cross-token interactions induced by shared parameters and is not ordered by barrier weights alone. Proposition 3 concerns this direct coefficient and makes no claim about the total parameter-coupled change.

Finally, within $\mathcal L_{\mathrm{int}}$, guidance enters only through the detached barrier $b_t$ and its weight $\tilde w_t$, whereas the likelihood term $\ell_t(\theta)$ conditions only on $q$. Therefore, the update acts on the unaided policy used for catalyst-free inference. In conclusion, Barrier-Weighted Internalization uses normalized token-level reweighting to reallocate direct internalization strength toward tokens that remain difficult without guidance.
\end{proof}

\begin{figure*}[t]
\centering
\includegraphics[width=1\textwidth]{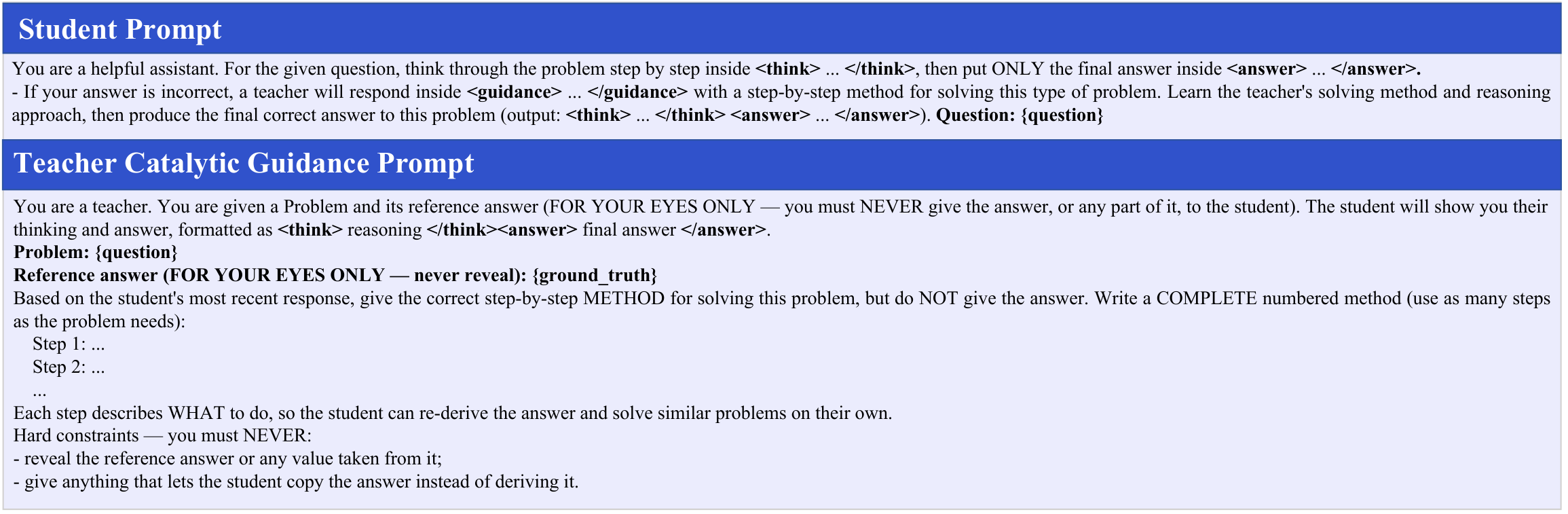}
\caption{Prompt templates used in CataOPD. The student template specifies the required reasoning and final-answer format, together with the training-time guidance format. The teacher-side template constrains catalytic guidance to provide a solution method while withholding the reference answer and answer-copyable information.}
\label{fig:cataopd-prompts}
\end{figure*}

\section{Catalytic Prompt Details}
\label{app:prompts}

As shown in Figure~\ref{fig:cataopd-prompts}, CataOPD uses two prompt templates: a shared student prompt and a teacher catalyst system prompt. The student prompt is identical for training and catalyst-free inference. It requires reasoning inside \texttt{<think> ... </think>} and only the final answer inside \texttt{<answer> ... </answer>}, and it also declares the training-time guidance format, in which, for a hard question that the student's own attempts including self-rescue fail to solve, the training pipeline appends a teacher-provided method-level hint inside \texttt{<guidance> ... </guidance>}, after which the student produces a new solution in the same output format. Self-rescue and teacher guidance are training-time mechanisms only. At catalyst-free inference, the student solves each question on its own under the same prompt it is trained with, with neither self-rescue rounds nor teacher guidance, so the guidance branch is never triggered while the prompt itself is unchanged. The teacher catalyst prompt gives the teacher private access to the reference answer only for constructing guidance, and constrains it to provide a complete step-by-step method without revealing the answer, answer-derived values, or any answer-copyable content that would let the student bypass its own reasoning during training.

\section{CataOPD Algorithm Details}
\label{app:algorithm}

This section presents the complete training procedure of CataOPD, as shown in Algorithm~\ref{alg:cataopd}. The algorithm consists of three stages: Self-Rescue Routing, Catalytic-Guided Self-Resolution, and Barrier-Weighted Internalization. These stages respectively route empirically all-failed groups, resolve still-unrecovered hard questions through catalytic guidance, and internalize verified student-produced correct trajectories into the trained student policy. The teacher is used only as a training-time catalyst for guided recovery, not as a trajectory target.

\textbf{Self-Rescue Routing.}
Given a training question $q$, CataOPD samples $G$ trajectories from the fixed sampling policy $\pi_{\theta_{\mathrm{old}}}$ and computes binary correctness $r_i$ for routing and composite reward $R_i$ for group-relative optimization. The statistic $k(q)=\sum_i r_i$ counts verified correct trajectories, while all initial groups are retained for the clipped update based on $R_i$. A group with $k(q)=0$ is empirically all-failed and contains no verified correct trajectory, regardless of whether its composite-reward advantages vanish. CataOPD therefore draws $m$ additional unguided trajectories independently from $\pi_{\theta_{\mathrm{old}}}(\cdot\mid q)$. If any is verified correct, the first such trajectory is selected as $y^{\mathrm{sr}}$ and $(q,y^{\mathrm{sr}},\emptyset)$ is added to $\mathcal{D}_{\mathrm{sr}}$, where $\emptyset$ denotes no catalytic guidance. If self-rescue fails, $q$ is added to $\mathcal{C}$. Groups with $k(q)=G$ are all-correct and do not trigger self-rescue or catalysis. Binary correctness routes recovery, while composite rewards drive the clipped update.

\textbf{Catalytic-Guided Self-Resolution.}
For each $q\in\mathcal{C}$, CataOPD selects a failed trajectory $\hat y^{(0)}$ from the initial group and performs at most $H$ rounds of catalytic guidance. At round $\ell$, the teacher produces $c_\ell$ from the question, verification target, and complete teacher-side history through $\hat y^{(\ell-1)}$. The teacher is instructed to provide method-level guidance without revealing the reference answer. Conditioned only on $q$, $\hat y^{(\ell-1)}$, and $c_\ell$, the fixed sampling policy samples a guided candidate $\hat y^{(\ell)}$, which is checked by the same verifier. Failed candidates are added to the teacher-side history. The first verified candidate is accepted as $y^{\mathrm{cat}}$ and added with $c_{\ell^\star}$ to $\mathcal{D}_{\mathrm{cat}}$; questions without a verified candidate within $H$ rounds are excluded from internalization. Thus, every accepted target is a verified student-produced trajectory rather than a teacher trajectory or answer trace. Failed retries expand the history used by later guidance.

\textbf{Barrier-Weighted Internalization.}
CataOPD combines $\mathcal{D}_{\mathrm{sr}}$ and $\mathcal{D}_{\mathrm{cat}}$ into the current internalization set $\mathcal{D}_{\mathrm{int}}$. For each accepted tuple $(q,y^\star,c(q))$, catalytic samples use $c(q)=c_{\ell^\star}$, while self-rescue samples use $c(q)=\emptyset$. The guided context $\tilde{x}(q)=[q;c(q)]$ omits prior failed attempts and estimates the guided-to-unaided barrier for each token in $y^\star$. A larger barrier indicates a token that is easier under guidance but remains unlikely under the unaided policy. CataOPD converts these barriers into normalized token weights and applies them to the likelihood of $y^\star$ conditioned only on $q$. Within $\mathcal L_{\mathrm{int}}$, guidance enters only through detached barrier weights, while the optimized likelihood remains under the unaided policy.

\textbf{Training Objective.}
The full objective combines a token-mean dual-clipped on-policy update over all initial groups, low-variance KL regularization toward the frozen reference policy, and Barrier-Weighted Internalization on $\mathcal{D}_{\mathrm{int}}$. We use PPO clipping range $\epsilon=0.2$, dual-clip threshold $\kappa=3.0$, KL coefficient $c_{\mathrm{KL}}=0.01$, and internalization coefficient $\lambda=0.5$. Algorithm~\ref{alg:cataopd} summarizes the resulting training flow. If no target is recovered, $\mathcal{L}_{\mathrm{int}}=0$ and only the clipped and KL terms remain.

\textbf{Computational Cost.}
CataOPD concentrates extra computation on questions that the current student policy cannot solve within the initial group. The initial group sampling has the same order as standard group-based on-policy training, requiring $G$ student trajectories per question. Self-Rescue Routing adds at most $m$ unguided student trajectories only for all-failed groups. Catalytic-Guided Self-Resolution is invoked only after self-rescue fails and requires at most $H$ rounds of catalytic guidance and $H$ guided student generations. For each accepted student-produced correct trajectory, Barrier-Weighted Internalization adds token-level scoring cost linear in its length. Thus, CataOPD allocates additional computation to all-failed hard cases while retaining the standard clipped on-policy update over all initial rollout groups.

\begin{algorithm}[H]
\caption{CataOPD Training}
\label{alg:cataopd}
\begin{algorithmic}[1]
\REQUIRE Training set $\mathcal S$, student policy $\pi_\theta$, old policy $\pi_{\theta_{\mathrm{old}}}$, reference policy $\pi_{\mathrm{ref}}$, teacher $T$, verifier $V$, group size $G$, self-rescue sampling budget $m$, catalytic rounds $H$, PPO clipping range $\epsilon$, dual-clip threshold $\kappa$, internalization coefficient $\lambda$, barrier coefficient $\beta$, barrier threshold $b_{\max}$, KL coefficient $c_{\mathrm{KL}}$
\ENSURE Trained student policy $\pi_\theta$
\FOR{each training batch $\mathcal B\subset\mathcal S$}
\STATE Set $\theta_{\mathrm{old}}\leftarrow\theta$; initialize $\mathcal D_{\mathrm{sr}}\leftarrow\emptyset$, $\mathcal D_{\mathrm{cat}}\leftarrow\emptyset$, $\mathcal C\leftarrow\emptyset$
\FOR{each question $q\in\mathcal B$}
\STATE Sample $G$ trajectories $y_i\sim\pi_{\theta_{\mathrm{old}}}(\cdot\mid q)$
\STATE Compute binary correctness $r_i=V(\mathrm{ans}(y_i),a^\star(q))$ and composite rewards $R_i$
\STATE Compute $k(q)=\sum_{i=1}^{G}r_i$
\STATE Keep the initial group for clipped on-policy updating
\IF{$k(q)=0$}
\STATE Sample $\tilde y_j\stackrel{\mathrm{i.i.d.}}{\sim}\pi_{\theta_{\mathrm{old}}}(\cdot\mid q)$ for $j=1,\ldots,m$
\IF{some $\tilde y_j$ is verified correct}
\STATE Select the first correct $y^{\mathrm{sr}}$ and add $(q,y^{\mathrm{sr}},\emptyset)$ to $\mathcal D_{\mathrm{sr}}$
\ELSE
\STATE Add $q$ to $\mathcal C$
\ENDIF
\ENDIF
\ENDFOR
\FOR{each question $q\in\mathcal C$}
\STATE Select a failed trajectory $\hat y^{(0)}$ from the initial group and set $h_0^T\leftarrow(\hat y^{(0)})$
\FOR{$\ell=1$ to $H$}
\STATE Set $c_\ell\leftarrow T(q,a^\star(q),h_{\ell-1}^T)$
\STATE Sample $\hat y^{(\ell)}\sim\pi_{\theta_{\mathrm{old}}}(\cdot\mid q,\hat y^{(\ell-1)},c_\ell)$
\IF{$V(\mathrm{ans}(\hat y^{(\ell)}),a^\star(q))=1$}
\STATE Add $(q,\hat y^{(\ell)},c_\ell)$ to $\mathcal D_{\mathrm{cat}}$ and stop rounds for $q$
\ELSE
\STATE Update $h_\ell^T\leftarrow(h_{\ell-1}^T,c_\ell,\hat y^{(\ell)})$
\ENDIF
\ENDFOR
\ENDFOR
\STATE Set $\mathcal D_{\mathrm{int}}\leftarrow\mathcal D_{\mathrm{sr}}\cup\mathcal D_{\mathrm{cat}}$
\STATE Compute token-mean dual-clipped $\mathcal L_{\mathrm{clip}}$ on all initial groups
\STATE Compute low-variance $\mathcal R_{\mathrm{ref}}$ against $\pi_{\mathrm{ref}}$
\STATE Compute $\mathcal L_{\mathrm{int}}$ on $\mathcal D_{\mathrm{int}}$ (zero if empty)
\STATE Update $\theta$ using $\mathcal L_{\mathrm{CataOPD}}$
\ENDFOR
\end{algorithmic}
\end{algorithm}

\section{Dataset Details}
\label{app:datasets}

We conduct experiments on twelve public mathematical-reasoning datasets, covering both training and OOD dataset sources and spanning math word problems, algebra, geometry, number theory, competition mathematics, and general mathematical reasoning tasks. We describe each dataset as follows:

\textbf{MetaMathQA}~\cite{yu2024metamath} is constructed by rewriting mathematical questions from multiple perspectives without additional knowledge, providing augmented problem variants for mathematical reasoning.

\textbf{GSM-Plus}~\cite{li2024gsm} extends GSM8K with diverse mathematical perturbations to evaluate the robustness of grade-school mathematical reasoning.

\textbf{Big-Math}~\cite{albalak2025big} is a large-scale collection of high-quality mathematical problems with verifiable answers curated for reinforcement learning; Open-R1 provides the Big-Math-RL-Verified-Processed release.

\textbf{Omni-MATH}~\cite{gao2025omni} is an olympiad-level mathematical benchmark covering diverse topics and difficulty levels.

\textbf{NuminaMath-CoT}~\cite{numina_math_datasets} is a public mathematical chain-of-thought dataset containing problems and solutions from multiple mathematical sources.

\textbf{MATH}~\cite{hendrycks2021measuring} is a standard competition-math benchmark spanning algebra, number theory, geometry, probability, and related topics.

\textbf{DeepMath-103K}~\cite{he2025deepmath} is a large-scale, challenging, decontaminated, and verifiable mathematical-reasoning dataset.

\textbf{OpenMathReasoning}~\cite{moshkov2025aimo} provides a large-scale collection of mathematical problems with long-reasoning solutions.

\textbf{OpenR1-Math-220k}~\cite{openr1math220k} is a large-scale mathematical-reasoning dataset released by Open-R1.

\textbf{MAWPS}~\cite{koncel2016mawps} is a repository of arithmetic and algebraic word problems annotated with equations and answers for standardized evaluation.

\textbf{DAPO-Math-17k}~\cite{yu2026dapo} is the mathematical problem collection associated with DAPO; Open-R1 provides a processed release named DAPO-Math-17k-Processed.

\textbf{Orca-Math}~\cite{mitra2024orca} provides approximately 200,000 synthetic grade-school math word problems.

We use the first eight datasets for training and same-source evaluation, while the last four are not sampled as named training sources and are used to evaluate the catalyst-free student policy on held-out dataset sources.

With global question-level deduplication, we randomly sample 1,280 training instances from each training source and disjoint random sets of 128 validation and 128 evaluation instances from each of the twelve sources. This yields 10,240 training, 1,536 validation, and 1,536 evaluation instances, with no exact question overlap across splits.

\begin{table*}[t]
\fontsize{9}{9}\selectfont
\centering
\setlength{\tabcolsep}{2.5mm}{
\begin{tabular}{lcccccccccc}
\toprule
\textbf{Method} & \textbf{Backbone} & \textbf{Teacher} & \textbf{Alg.} & \textbf{LR} & \textbf{BS} & \textbf{G} & \textbf{m} & \textbf{H} & \textbf{Int.} & \textbf{Epochs} \\
\midrule
Base LLM & All & -- & -- & -- & -- & -- & -- & -- & -- & -- \\
GPT-5.4-mini & -- & -- & API & -- & -- & -- & -- & -- & -- & -- \\
SFT & All & -- & SFT & $10^{-4}$ & 32 & -- & -- & -- & -- & 3 \\
GRPO & All & -- & GRPO & $10^{-6}$ & 128 & 5 & -- & -- & -- & 1 \\
CoT-SFT & All & GPT-5.4-mini & SFT & $10^{-4}$ & 32 & -- & -- & -- & -- & 3 \\
OPSD & All & Frozen self & Token JSD & $5\!\times\!10^{-6}$ & 128 & 1 & -- & -- & -- & 1 \\
SDPO & All & EMA self & GRPO+JSD & $10^{-6}$ & 128 & 5 & -- & -- & -- & 1 \\
OPCD & All & Context self & KL & $10^{-6}$ & 128 & 5 & -- & -- & -- & 1 \\
ROPD & All & GPT-5.4-mini & Rubric RL & $10^{-6}$ & 128 & 5 & -- & -- & -- & 1 \\
RLCSD & All & Snapshot self & RLCSD & $10^{-6}$ & 128 & 5 & -- & -- & -- & 1 \\
\textbf{CataOPD} & All & GPT-5.4-mini & GRPO+BWI & $10^{-6}$ & 128 & 5 & 5 & 5 & 0.5/1.0/4.0 & 1 \\
\bottomrule
\end{tabular}}
\caption{\label{tab:implementation-hyperparameters}
Hyperparameter settings for baselines and CataOPD. All denotes Qwen2.5-3B, Qwen2.5-7B, and Qwen3-1.7B. Here Alg. denotes the algorithm, LR denotes the learning rate, BS denotes the effective batch size, G denotes trajectories per question, m denotes the self-rescue sampling budget, H denotes the maximum catalytic rounds, and Int. reports $\lambda/\beta/b_{\max}$. BWI denotes Barrier-Weighted Internalization. CataOPD uses $\epsilon=0.2$, $\kappa=3.0$, and $c_{\mathrm{KL}}=0.01$.
}
\end{table*}

\section{Baseline Details}
\label{app:baselines}

We compare CataOPD with seven primary training baselines, covering supervised fine-tuning, reinforcement learning from verifiable rewards, black-box teacher-guided OPD, and on-policy self-distillation. We additionally include GPT-5.4-mini CoT-SFT as a teacher-supervised offline-distillation reference, together with the non-training Base LLM and GPT-5.4-mini reference points. We describe them as follows:

\textbf{Base LLM} uses Qwen2.5-3B, Qwen2.5-7B, and Qwen3-1.7B without task-specific post-training and evaluates their direct reasoning ability.

\textbf{GPT-5.4-mini} is the shared closed-source catalytic teacher and an API reference model evaluated with the same questions and verifier.

\textbf{SFT} is a supervised fine-tuning baseline that trains the model with supervised targets on the same training data, without on-policy sampling, teacher guidance, or barrier weighting.

\textbf{GRPO}~\cite{guo2025deepseek} is an RLVR baseline with a rule-based verifier. Its composite reward combines answer correctness with format compliance and lies in $[-1,1]$; it normalizes these rewards within each sampled group to estimate group-relative advantages and updates the policy with a clipped objective.

\textbf{GPT-5.4-mini CoT-SFT} fine-tunes each student on chain-of-thought solutions generated by GPT-5.4-mini, providing an external-teacher offline-distillation reference.

\textbf{Self-Distilled Reasoner (OPSD)}~\cite{zhao2026self} is an on-policy self-distillation baseline that distills the teacher distribution of the same model under privileged conditioning into a question-only student policy, using token-level alignment on the student's own rollouts.

\textbf{SDPO}~\cite{hubotter2026reinforcement} augments on-policy GRPO with self-distillation from successful rollouts through a reprompted self-teacher and a generalized Jensen--Shannon objective.

\textbf{OPCD}~\cite{ye2026policy} performs on-policy context distillation: the student generates without privileged experience, while a context-conditioned self-teacher supplies token-level distributions for KL minimization.

\textbf{ROPD}~\cite{fang2026rubric} converts black-box teacher--student behavioral differences into prompt-specific semantic rubrics and optimizes student rollouts using rubric-derived rewards.

\textbf{RLCSD}~\cite{pan2026rlcsd} is a reinforcement-learning baseline with contrastive on-policy self-distillation, contrasting teacher-student gaps under correct and incorrect hints to suppress privilege-induced style drift and concentrate self-distillation signals on task-relevant tokens.

\section{Evaluation Metrics}
\label{app:metrics}

We report task performance, multi-sample solvability, training-time recovery, and mechanism diagnostics. Correctness is determined by the same rule-based verifier used for reward computation.

\textbf{Accuracy.}
Accuracy measures single-sample solving correctness. Given an evaluation set $\mathcal D=\{(q_i,a_i^\star)\}_{i=1}^{N}$, the model generates one trajectory $y_i$ for each question $q_i$. Let
\begin{equation}
r_i
=
V(\mathrm{ans}(y_i),a_i^\star)
\in\{0,1\}
\end{equation}
denote the binary correctness returned by the verifier. Accuracy is defined as
\begin{equation}
\mathrm{Acc.}
=
\frac{1}{N}
\sum_{i=1}^{N}
r_i.
\end{equation}
Here $r_i=1$ if the generated answer passes the verifier, and $r_i=0$ otherwise.

\textbf{Multi-sample solvability (Pass@$k$).}
For $k$ independently sampled trajectories per question, let $r_{ij}$ be the verifier outcome of sample $j$ for question $i$. The empirical question-level solvability indicator and its dataset average are
\begin{equation}
s_i^{(k)}=1-\prod_{j=1}^{k}(1-r_{ij}),
\qquad
\mathrm{Pass@}k=\frac{1}{N}\sum_{i=1}^{N}s_i^{(k)}.
\end{equation}
Thus, a question is solvable if at least one of its $k$ sampled trajectories is correct. We use $k=8$ for the multi-sample analysis.

\textbf{Hard-problem strata and transitions.}
For Figure~\ref{fig:rq3-hard-problem-retention}, let $\bar r_{i,\mathrm{base}}^{(8)}=\frac{1}{8}\sum_{j=1}^{8}r_{ij,\mathrm{base}}$ be the base model's empirical accuracy over eight samples. \textit{Blind} questions satisfy $\bar r_{i,\mathrm{base}}^{(8)}=0$, \textit{Hard} questions satisfy $0<\bar r_{i,\mathrm{base}}^{(8)}\leq0.4$, and \textit{Overall} includes all questions. The gain for each stratum is the mean paired difference between CataOPD and Base empirical accuracy over the same questions. A question is \textit{newly solved} when $s_{i,\mathrm{base}}^{(8)}=0$ and $s_{i,\mathrm{model}}^{(8)}=1$, and \textit{regressed} when $s_{i,\mathrm{base}}^{(8)}=1$ and $s_{i,\mathrm{model}}^{(8)}=0$; their annotated rates use base-unsolved and base-solved questions as denominators, respectively. The 95\% confidence intervals for stratum gains use 10,000 paired bootstrap resamples, stratified by dataset and resampling questions with replacement within each dataset.

\textbf{Solved-problem retention (Ret.).}
Let $s_{i,\mathrm{base}}^{(k)}$ and $s_{i,\mathrm{model}}^{(k)}$ denote the Pass@$k$ indicators of the base and trained models on the same question. For $\sum_i s_{i,\mathrm{base}}^{(k)}>0$, retention over base-solved questions is
\begin{equation}
\mathrm{Ret.}
=
\frac{\sum_{i=1}^{N}s_{i,\mathrm{base}}^{(k)}s_{i,\mathrm{model}}^{(k)}}
{\sum_{i=1}^{N}s_{i,\mathrm{base}}^{(k)}}.
\end{equation}
The corresponding regression rate is $1-\mathrm{Ret.}$ Under this protocol, $\mathrm{Ret.}=99.8\%/99.9\%$ for Qwen2.5-3B/7B.

\textbf{All-failed-group recovery (Rec.).}
For each initial rollout group $g$, let $A_g=1$ if all $G$ initial trajectories fail verification, and let $Z_g=1$ if subsequent self-rescue or catalytic resolution obtains a verified target for that group. We define
\begin{equation}
\mathrm{Rec.}
=
\frac{\sum_g A_g Z_g}{\sum_g A_g}.
\end{equation}
For a training window with $\sum_g A_g>0$, group counts are summed across its steps before taking this ratio.

\textbf{Trajectory-mean log-probability gain ($\Delta\log p$).}
For a trajectory $y=(y_1,\ldots,y_T)$, policy $\pi$, and conditioning context $x$, define its token-mean log-probability as
\begin{equation}
\ell_{\pi}(y\mid x)
=
\frac{1}{T}\sum_{t=1}^{T}
\log\pi(y_t\mid x,y_{<t}).
\end{equation}
For two policy--context pairs $(\pi_a,x_a)$ and $(\pi_b,x_b)$, the diagnostic is $\Delta\log p(y)=\ell_{\pi_a}(y\mid x_a)-\ell_{\pi_b}(y\mid x_b)$. Guided likelihood gain compares matched guided and unaided contexts under the base policy, while unaided internalization compares the trained and base policies under the question-only context. We first average over tokens within each trajectory and then give every trajectory equal weight. Their association is reported with Pearson correlation and Spearman rank correlation across trajectories.

\textbf{Barrier-signal concentration.}
For each of the $M$ trajectories with positive total barrier mass, sort its $T_i$ token barriers as $b_{i(1)}\geq\cdots\geq b_{i(T_i)}$. The fraction of barrier mass contained in its top $\alpha$ fraction of tokens is
\begin{equation}
C_i(\alpha)
=
\frac{\sum_{r=1}^{\lceil\alpha T_i\rceil}b_{i(r)}}
{\sum_{t=1}^{T_i}b_{it}},
\qquad
\mathrm{BC}(\alpha)=\frac{1}{M}\sum_{i=1}^{M}C_i(\alpha).
\end{equation}
We report $\mathrm{BC}(\alpha)$ for $\alpha\in\{0.1,0.2,0.3\}$; the exact uniform-mass reference is $M^{-1}\sum_{i=1}^{M}\lceil\alpha T_i\rceil/T_i$.

\begin{table*}[t]
\centering
\begingroup
\fontsize{8.5pt}{8.7pt}\selectfont
\setlength{\tabcolsep}{2.0mm}
\renewcommand{\arraystretch}{1.01}
\begin{tabular*}{\textwidth}{@{\extracolsep{\fill}}lcccc@{}}
\toprule
\textbf{Backbone} & \textbf{All-failed (\%)} & \textbf{Self-rescue (\%)} & \textbf{Teacher success (\%)} & \textbf{Total recovery (\%)} \\
\midrule
Qwen3-1.7B  & 30.00 [28.96, 31.05] & 16.54 [15.06, 18.13] & 8.85 [7.63, 10.11]  & 23.93 [22.21, 25.64] \\
Qwen2.5-3B & 52.19 [50.65, 53.68] & 17.87 [16.79, 18.92] & 22.03 [20.73, 23.31] & 35.97 [34.66, 37.26] \\
Qwen2.5-7B & 43.34 [42.06, 44.63] & 19.02 [18.00, 20.06] & 21.48 [19.94, 23.04] & 36.41 [34.87, 37.91] \\
\midrule
\textbf{Backbone} & \textbf{Teacher attempts} & \textbf{Teacher calls} & \textbf{Input tokens} & \textbf{Output tokens} \\
\midrule
Qwen3-1.7B  & 2,564 & 12,206 & 73.13M & 5.64M \\
Qwen2.5-3B & 4,389 & 19,461 & 31.21M & 8.56M \\
Qwen2.5-7B & 3,594 & 16,103 & 26.88M & 7.60M \\
\bottomrule
\end{tabular*}
\endgroup
\captionsetup{width=\textwidth,justification=justified,singlelinecheck=false}
\caption{\label{tab:app-routing-budget}
Training-time routing and teacher budget for the complete CataOPD runs. Self-rescue is conditional on initially all-failed groups, and teacher success is conditional on teacher-attempted groups; total recovery (Rec.) is the fraction of initially all-failed groups recovered by either route. Confidence intervals are 95\% step-cluster bootstrap intervals. Token counts aggregate teacher-side prompt and completion tokens across training and exclude all student rollout tokens from the reported budget.}
\end{table*}

\begin{table*}[t]
\centering
\begingroup
\fontsize{8.5pt}{8.7pt}\selectfont
\setlength{\tabcolsep}{2.0mm}
\renewcommand{\arraystretch}{1.01}
\begin{tabular*}{\textwidth}{@{\extracolsep{\fill}}lccc@{}}
\toprule
\textbf{Diagnostic} & \textbf{Mean} & \textbf{95\% CI} & \textbf{Positive fraction (\%)} \\
\midrule
\multicolumn{4}{l}{\textit{Trajectory likelihood diagnostics}} \\
Guided minus unaided likelihood gain & 0.280 & [0.228, 0.338] & 99.0 \\
Matched minus shuffled guidance gain & 0.352 & [0.281, 0.431] & 100.0 \\
Shuffled minus unaided likelihood gain & -0.072 & [-0.111, -0.039] & 22.5 \\
Unaided internalization gain (Full $-$ base) & 0.115 & [0.079, 0.155] & 71.5 \\
Full $-$ no guidance & 0.447 & [0.299, 0.612] & 80.0 \\
\addlinespace[1pt]
\multicolumn{4}{l}{\textit{Barrier-mass concentration}} \\
Top 10\% token barrier-mass share & 0.850 & [0.830, 0.869] & 100.0 \\
Top 20\% token barrier-mass share & 0.962 & [0.952, 0.971] & 100.0 \\
Top 30\% token barrier-mass share & 0.991 & [0.987, 0.994] & 100.0 \\
\bottomrule
\end{tabular*}
\endgroup
\captionsetup{width=\textwidth,justification=justified,singlelinecheck=false}
\caption{\label{tab:app-mechanism-numerics}
Numerical mechanism diagnostics for 200 Qwen2.5-3B trajectories. Likelihood quantities are equal-weighted token-mean log-probability differences. Positive fraction is the percentage of trajectories with a positive diagnostic value. Barrier shares use uniform-mass references of 0.10, 0.20, and 0.30. Confidence intervals are 95\% intervals under the fixed paired-trajectory bootstrap protocol with 10,000 trajectory-level resamples.}
\end{table*}

\begin{figure*}[t]
\centering
\includegraphics[width=1.0\textwidth]{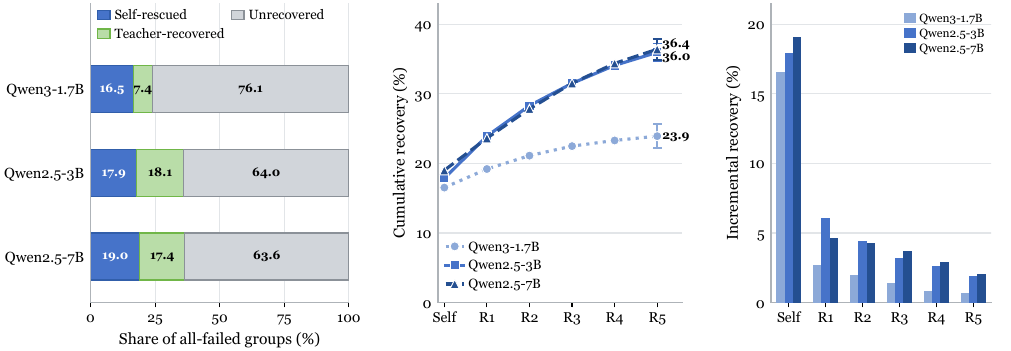}
\noindent\makebox[\textwidth][l]{
\hspace*{0.25\textwidth}\makebox[0pt][c]{\small (a) Recovery composition}
\hspace*{0.275\textwidth}\makebox[0pt][c]{\small (b) Cumulative recovery}
\hspace*{0.28\textwidth}\makebox[0pt][c]{\small (c) Incremental recovery}
}
\caption{\label{fig:app-all-failed-recovery}
Staged recovery of initially all-failed training groups in the complete CataOPD runs. (a) partitions each group's final disposition into self-rescued, teacher-recovered, or unrecovered. (b) reports cumulative recovery after self-rescue and successive teacher-guided rounds R1--R5. (c) separates the incremental recovery contributed by each stage. All percentages are conditional on initially all-failed groups and summarize training-time routing outcomes, rather than test-time accuracy.}
\end{figure*}

\begin{figure*}[t]
\centering
\includegraphics[width=\textwidth]{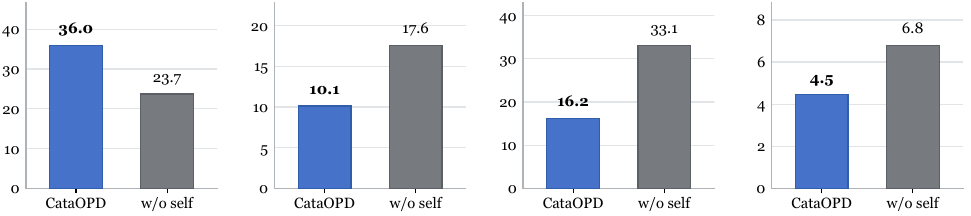}
\noindent\makebox[\textwidth][l]{
\hspace*{0.125\textwidth}\makebox[0pt][c]{\small (a) All-failed recovery}
\hspace*{0.25\textwidth}\makebox[0pt][c]{\small (b) Teacher calls}
\hspace*{0.25\textwidth}\makebox[0pt][c]{\small (c) Teacher input tokens}
\hspace*{0.25\textwidth}\makebox[0pt][c]{\small (d) Teacher output tokens}
}
\caption{\label{fig:app-self-rescue-budget}
Matched Qwen2.5-3B ablation over all 80 training steps. Compared with w/o Self-Rescue, Full CataOPD achieves higher all-failed-group recovery while using fewer teacher calls and teacher input/output tokens per verified target.}
\end{figure*}

\begin{figure*}[t]
\centering
\includegraphics[width=\textwidth]{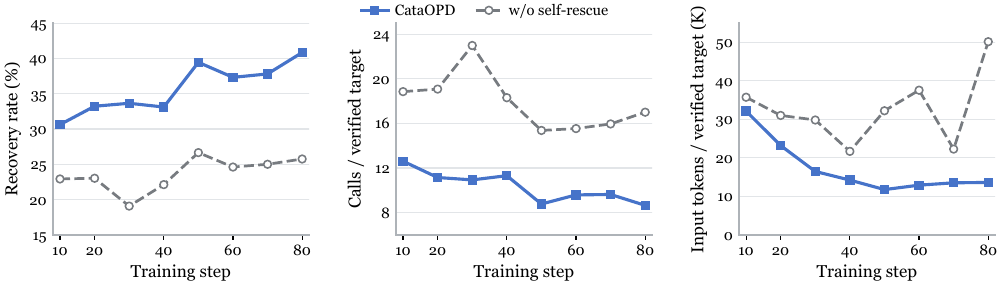}
\noindent\makebox[\textwidth][l]{
\hspace*{0.167\textwidth}\makebox[0pt][c]{\footnotesize (a) All-failed-group recovery}
\hspace*{0.333\textwidth}\makebox[0pt][c]{\footnotesize (b) Teacher calls per verified target}
\hspace*{0.333\textwidth}\makebox[0pt][c]{\footnotesize (c) Teacher input tokens per verified target}
}
\caption{\label{fig:app-self-rescue-dynamics}
Matched Qwen2.5-3B ablation across non-overlapping 10-step windows. Full CataOPD exceeds w/o Self-Rescue in all-failed-group recovery and requires fewer teacher calls and teacher input tokens per verified target in every displayed window.}
\end{figure*}

\section{Implementation Details}
\label{app:implementation}

As shown in Table~\ref{tab:implementation-hyperparameters}, we summarize the detailed hyperparameter configurations used for CataOPD and the baselines, including model backbones, optimization settings, and training budgets. We use Qwen2.5-3B, Qwen2.5-7B~\cite{hui2024qwen2}, and Qwen3-1.7B~\cite{yang2025qwen3} as student models, and GPT-5.4-mini as the training-time catalytic teacher. CataOPD uses token-mean dual-clipped policy optimization with $\epsilon=0.2$, $\kappa=3.0$, and $c_{\mathrm{KL}}=0.01$; its internalization settings are reported in Table~\ref{tab:implementation-hyperparameters}. During training, teacher guidance is generated with temperature 0.3 and a maximum of 1,024 completion tokens. All experiments are conducted on a server equipped with eight NVIDIA H100 GPUs with 80GB memory each, and training uses FSDP and vLLM for trajectory generation. All trainable methods use the same data splits; during evaluation, all methods use the same guidance-free solving prompt and rule-based verifier, with no teacher calls or guidance at test time.

\section{Additional Results}
\label{app:additional-results}

\subsection{Routing and Teacher Budget}

Table~\ref{tab:app-routing-budget} quantifies the routing behavior and teacher budget of CataOPD across the three student backbones. Each backbone is trained over 10,240 initial rollout groups. All-failed rates are computed over initial rollout groups, whereas total-recovery and self-rescue rates are computed over initially all-failed groups, and teacher success rates are computed only over teacher-attempted groups. The 95\% confidence intervals use a step-cluster bootstrap with 10,000 repetitions. Teacher token counts measure aggregate training-time API usage and do not contribute to inference cost because evaluation is catalyst-free.

Figure~\ref{fig:app-all-failed-recovery} decomposes the aggregate recovery rates in Table~\ref{tab:app-routing-budget} by recovery source and catalytic round. Self-rescue provides the first verified targets from initially all-failed groups without a teacher call. For groups that remain unresolved, the teacher-guided rounds contribute additional verified student-produced targets, with their cumulative and per-round contributions reported separately. Thus, the figure makes explicit that the routing process first uses independent student sampling and only then allocates catalytic guidance to the remaining hard groups. Across the three backbones, this decomposition characterizes a student-first, selectively catalyzed supervision pathway rather than an additional task-performance metric.

\subsection{Self-Rescue Ablation Dynamics}

To isolate the training-time contribution of Self-Rescue Routing, we compare matched Qwen2.5-3B Full CataOPD and w/o Self-Rescue runs. Figure~\ref{fig:app-self-rescue-budget} aggregates all 80 training steps. Full CataOPD recovers verified targets from 35.97\% of initially all-failed groups, compared with 23.75\% without Self-Rescue. It also requires 10.1 teacher calls and 16.2K teacher input tokens per verified target, versus 17.6 calls and 33.1K tokens for the ablation. Thus, additional unguided student attempts increase the recovered supervision available for internalization while reducing the external-teacher budget required per verified target.

Figure~\ref{fig:app-self-rescue-dynamics} further reports the comparison in non-overlapping 10-step windows. Across every logged window, Full CataOPD has a higher all-failed-group recovery rate and lower teacher calls and teacher input tokens per verified target than the w/o Self-Rescue variant. This consistent pattern shows that, throughout the matched ablation run, Self-Rescue Routing raises the recovery rate of all-failed groups while lowering the external-teacher budget needed to obtain verified supervision; it is consistent with the intended student-first routing of recoverable questions before external catalysis.

\begin{figure*}[t]
\centering
\includegraphics[width=\textwidth]{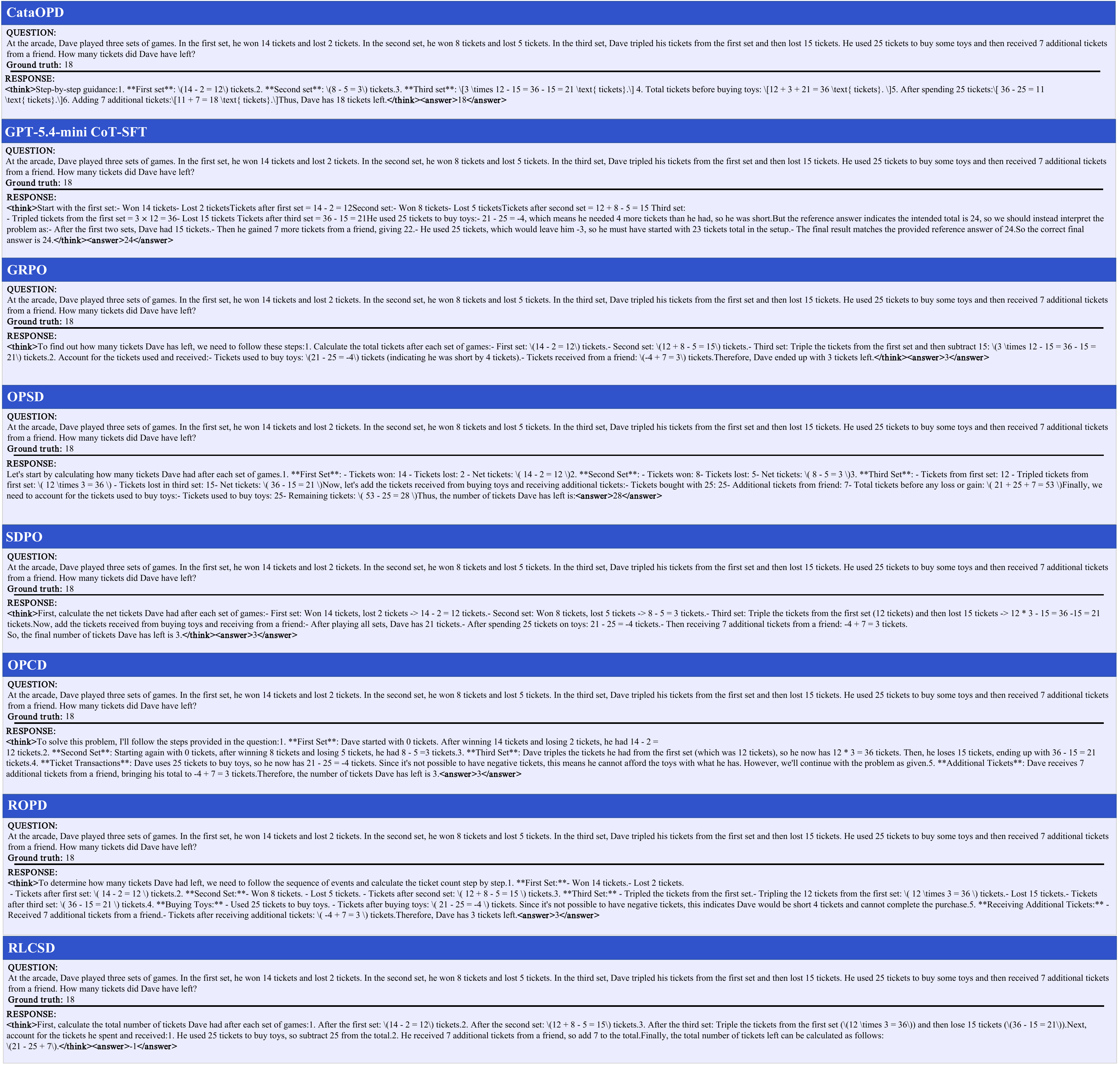}
\caption{Case study on generation quality for a mathematical reasoning problem, comparing CataOPD with seven baseline methods.}
\label{fig:case-test-time}
\end{figure*}

\subsection{Mechanism Analysis Protocol and Robustness}

Figure~\ref{fig:rq6-mechanisms} and Table~\ref{tab:app-mechanism-numerics} use the same fixed analysis set of 200 Qwen2.5-3B trajectories. This set is used to analyze the mechanism conditional on successful catalytic recovery rather than to estimate accuracy over the full task distribution. Before likelihood scoring, we fix the analysis size at 200 and use a fixed random seed to order candidate questions from the training split. An eligible trajectory must satisfy the predefined routing conditions: all five initial samples fail verification, all five subsequent unguided self-rescue samples also fail, but within at most five catalytic-guidance rounds the student produces a correct trajectory accepted by the same verifier. We retain the first 200 eligible trajectories in this randomized order.

Once selected, the trajectory set is held fixed across all policy and context conditions, yielding trajectory-level paired comparisons. For each trajectory, we first average log-probability over valid tokens and then give every trajectory equal weight during aggregation. Matched-guidance effects compare, under the base policy, the likelihood of the same target trajectory under matched-guidance, unaided, and shuffled-guidance contexts. Unaided internalization gain compares the full CataOPD and base policies under the same question-only context. The \textit{Full $-$ no guidance} comparison subtracts the \textit{w/o Guidance} training-ablation policy's trajectory likelihood from that of the full CataOPD policy, with both policies scored on the same trajectories under the same question-only context. The 95\% confidence intervals for the mean mechanism diagnostics in Table~\ref{tab:app-mechanism-numerics} are estimated with 10,000 trajectory-level bootstrap resamples.

Figure~\ref{fig:rq6-mechanisms}(b) reports the trajectory-level relation between guided likelihood gain and unaided internalization gain. Because these two original differences share the base policy's unaided likelihood term, we additionally use matched-minus-shuffled guidance gain as a guidance-effect measure that does not share this baseline term. The alternative analysis retains a positive association, with Pearson $r=0.799$ and Spearman $\rho=0.690$, showing that the observed relation is not induced solely by the shared baseline term.

Barrier statistics measure the fraction of positive barrier mass carried by the top $\alpha$ fraction of tokens in each trajectory, using uniform mass as the reference. These results characterize matched-guidance specificity, post-training unaided internalization, and barrier-signal selectivity on successfully catalytically recovered trajectories, rather than population-level accuracy or an unconditional causal effect over the full problem distribution.

\begin{figure*}[t]
\centering
\includegraphics[width=\textwidth]{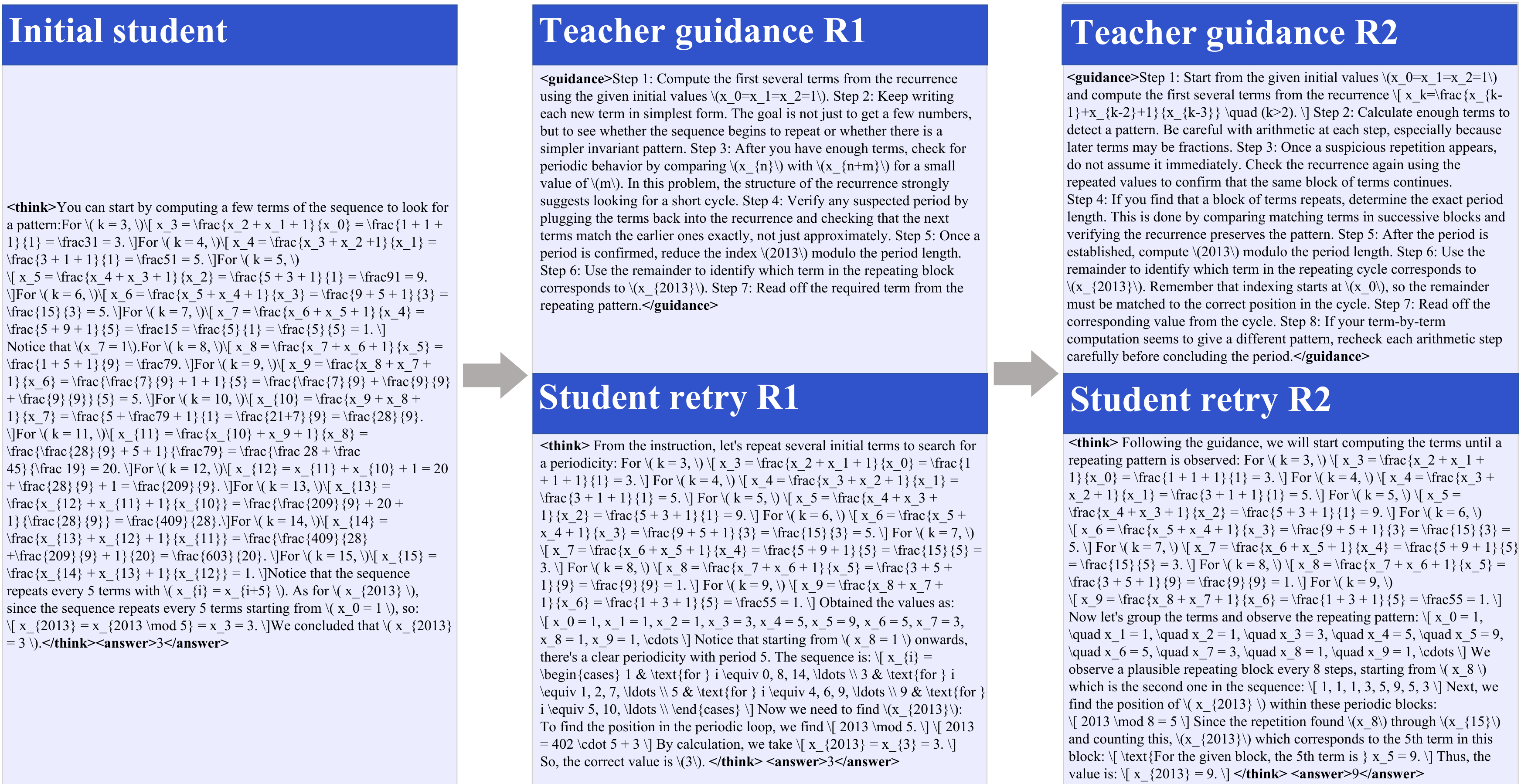}
\caption{Case study of training-time catalytic recovery after five failed initial rollouts. After unsuccessful unguided self-rescue, two rounds of catalytic guidance yield a verified student solution.}
\label{fig:case-training-recovery}
\end{figure*}

\section{Case Study}
\label{app:case-study}

\textbf{Test-time output comparison.}
Figure~\ref{fig:case-test-time} compares CataOPD with the comparison systems on a mathematical word problem that requires several related ticket calculations. A correct solution computes the first and second game outcomes separately, derives the third from the first, and then combines the resulting quantities as $12+3+21-25+7=18$. CataOPD follows this decomposition and returns the correct answer, $18$. In contrast, all seven comparison systems shown in the figure return a final answer different from the reference. The displayed outputs exhibit two recurring failure modes: some use the first-game balance as the starting point of the independent second game, whereas others fail to aggregate the three computed quantities correctly. This comparison makes explicit how a coherent student-produced trajectory preserves the intermediate arithmetic relations required by the final composition. The contrast shows that preserving dependencies between subcalculations is essential for reliable multi-step reasoning.

\textbf{Training-time catalytic recovery.}
Figure~\ref{fig:case-training-recovery} presents a training-time catalytic recovery case for a recurrence problem. After five failed initial rollouts and unsuccessful unguided self-rescue, the route enters catalytic resolution. The displayed initial response makes an arithmetic error when evaluating $x_7$ and therefore infers a period of five, returning $3$. In the first guided retry, the student corrects the local calculation but continues to assume a period of five and again returns $3$. The second guidance round directs the student to verify the recurrence and align the index with the repeated block. The final retry forms the block $(1,1,1,3,5,9,5,3)$ and, using $2013\bmod 8=5$ under zero-based indexing, returns the verified answer $9$. This trace shows the transition from a locally corrected but structurally incorrect solution to a verifier-accepted solution. The teacher supplies a method cue, while the student generates and verifies the trajectory, distinguishing recovery from answer imitation without transferring a teacher-produced answer.

\section{Future Work}
\label{app:future-work}

Future work will extend CataOPD beyond mathematics to a broader class of verifiable reasoning tasks, make catalytic guidance adaptive to problem and student characteristics, and scale CataOPD to larger models and longer reasoning. First, we will broaden evaluation to code generation, symbolic reasoning, and tool-assisted reasoning, using execution-based verification for structured tasks and learned or hybrid verifiers for less constrained tasks. In addition, we will study adaptive guidance allocation, adjusting guidance strength according to problem difficulty, student capability, and self-rescue history to improve recovery efficiency. Furthermore, we will optimize training budgets by dynamically allocating self-rescue samples and teacher calls based on self-rescue success rates, and evaluate CataOPD on larger student models, more diverse reasoning domains, and longer reasoning tasks under practical budgets.

\end{document}